\pdfoutput=1

\documentclass[11pt]{article}

\usepackage[final]{acl}

\usepackage{fancyhdr}
\usepackage{times}
\usepackage{latexsym}

\usepackage[T1]{fontenc}

\usepackage[utf8]{inputenc}

\usepackage{microtype}

\usepackage{inconsolata}

\usepackage{graphicx}

\usepackage{booktabs}       
\usepackage{amsfonts}       
\usepackage{nicefrac}       
\usepackage{microtype}      
\usepackage{xcolor}         
\usepackage{multirow}
\usepackage{xspace}
\usepackage{soul}

\usepackage{longtable}
\usepackage{ulem}
\usepackage{multirow}
\usepackage{siunitx}
\usepackage{float}
\usepackage{array}
\usepackage{amsmath}
\usepackage{subcaption}
\usepackage{enumitem}
\usepackage{xcolor}
\usepackage{xspace}
\usepackage{booktabs}
\usepackage{tabularx}
\usepackage[utf8]{inputenc}
\usepackage{geometry}
\usepackage{tcolorbox}
\usepackage{lipsum}
\usepackage{makecell}
\usepackage{enumitem}
\usepackage{dcolumn}
\newcolumntype{d}[1]{D{.}{.}{#1}}

\newcommand*\samethanks[1][\value{footnote}]{\footnotemark[#1]}
 
\definecolor{darkgreen}{RGB}{0,100,0}
\definecolor{red}{RGB}{255,0,0}
\newcommand{\eval}[1]{{\fontfamily{phv}\selectfont #1}}
\newcommand{\judge}[1]{\texttt{\textcolor{black}{#1}}}

\newcommand{\njudgesword}{thirteen\xspace}
\newcommand{\nexamtakers}{9\xspace}
\newcommand{\nexamtakersword}{nine\xspace}

\newcommand{\evaluatormodel}{exam-taker model\xspace}
\newcommand{\evaluatormodels}{exam-taker models\xspace}
\newcommand{\Evaluatormodel}{Exam-taker model\xspace}
\newcommand{\Evaluatormodels}{Exam-taker models\xspace}

\newcommand{\judgemodel}{judge model\xspace}
\newcommand{\judgemodels}{judge models\xspace}
\newcommand{\Judgemodel}{Judge model\xspace}

\newcommand{\Judgemodels}{Judge models\xspace}

\newcommand{\scottspi}{Scott's $\mathbf{\pi}$\xspace}

\newcommand{\gpt}{GPT-4\;Turbo\xspace}

\newcommand{\specialcell}[2][c]{\footnotesize\begin{tabular}[#1]{@{}l@{}}#2\end{tabular}}

\tcbset{
  outerbox/.style={
    colback=white,
    colframe=black,
    fonttitle=\bfseries,
    boxrule=0.25mm,
    sharp corners,
    rounded corners,
  }
}

\def\mytcolorbox#1#2#3{
  \begin{tcolorbox}[outerbox, title={\textit{#1}}]
  \texttt{#2} \\
  \\
  \texttt{#3}
  \end{tcolorbox}
}

\usepackage[noabbrev,capitalize,nameinlink]{cleveref}
\crefname{section}{Section}{Sections}
\crefname{table}{Table}{}
\crefname{figure}{Figure}{}
\crefname{section}{\S}{\S\S}
\Crefname{section}{\S}{\S\S}
\crefname{appendix}{Appendix}{Appendices}
\Crefname{Appendix}{Appendix}{}

\title{Judging the Judges: Evaluating Alignment and Vulnerabilities in LLMs-as-Judges}

\author{Aman Singh Thakur\thanks{Equal Contribution} \and Kartik Choudhary\samethanks \and Venkat Srinik Ramayapally\samethanks \\
  University of Massachusetts Amherst \\
  \texttt{\{amansinghtha, kartikchoudh, vramayapally\}@umass.edu} \\\AND
  Sankaran Vaidyanathan \\
  University of Massachusetts Amherst \\
  \texttt{sankaranv@cs.umass.edu} \\\And  
  Dieuwke Hupkes \\
  Meta \\
  \texttt{dieuwkehupkes@meta.com}
  }

\begin{document}
\maketitle

\setcounter{page}{404}

\fancypagestyle{firstpage}{
  \fancyhf{} 
  \renewcommand{\headrulewidth}{0pt}
  \renewcommand{\footrulewidth}{0pt}
  \fancyfoot[C]{\thepage\\[4pt]
    \footnotesize
    \parbox{\textwidth}{\centering
      \textit{Proceedings of the Fourth Workshop on Generation, Evaluation and Metrics (GEM$^2$ 2025), pages 404--430}\\[2pt]
      July 31 -- August 1, 2025 \copyright~2025 Association for Computational Linguistics
    }
  }
}

\fancypagestyle{otherpages}{
  \fancyhf{}
  \renewcommand{\headrulewidth}{0pt}
  \renewcommand{\footrulewidth}{0pt}
  \fancyfoot[C]{\thepage} 
}

\thispagestyle{firstpage}
\pagestyle{otherpages}

\begin{abstract}

The \textit{LLM-as-a-judge} paradigm offers a potential solution to scalability issues in human evaluation of large language models (LLMs), but there are still many open questions about its strengths, weaknesses, and potential biases. This study investigates \njudgesword models, ranging in size and family, as `\textit{\judgemodels}' evaluating answers from \nexamtakersword base and instruction-tuned `\textit{\evaluatormodels}'. We find that only the best (and largest) models show reasonable alignment with humans, though they still differ with up to 5 points from human-assigned scores. Our research highlights the need for alignment metrics beyond percent agreement, as judges with high agreement can still assign vastly different scores. We also find that smaller models and the lexical metric \judge{contains} can provide a reasonable signal in ranking the \evaluatormodels. Further error analysis reveals vulnerabilities in \judgemodels, such as sensitivity to prompt complexity and a bias toward leniency. Our findings show that even the best \judgemodels differ from humans in this fairly sterile setup, indicating that caution is warranted when applying \judgemodels in more complex scenarios.

\end{abstract}

\section {Introduction} \label{sec:intro}

\begin{figure*}[t]
    \centering
    \begin{subfigure}[b]{0.55\textwidth}
        \centering
        \includegraphics[width=\linewidth]{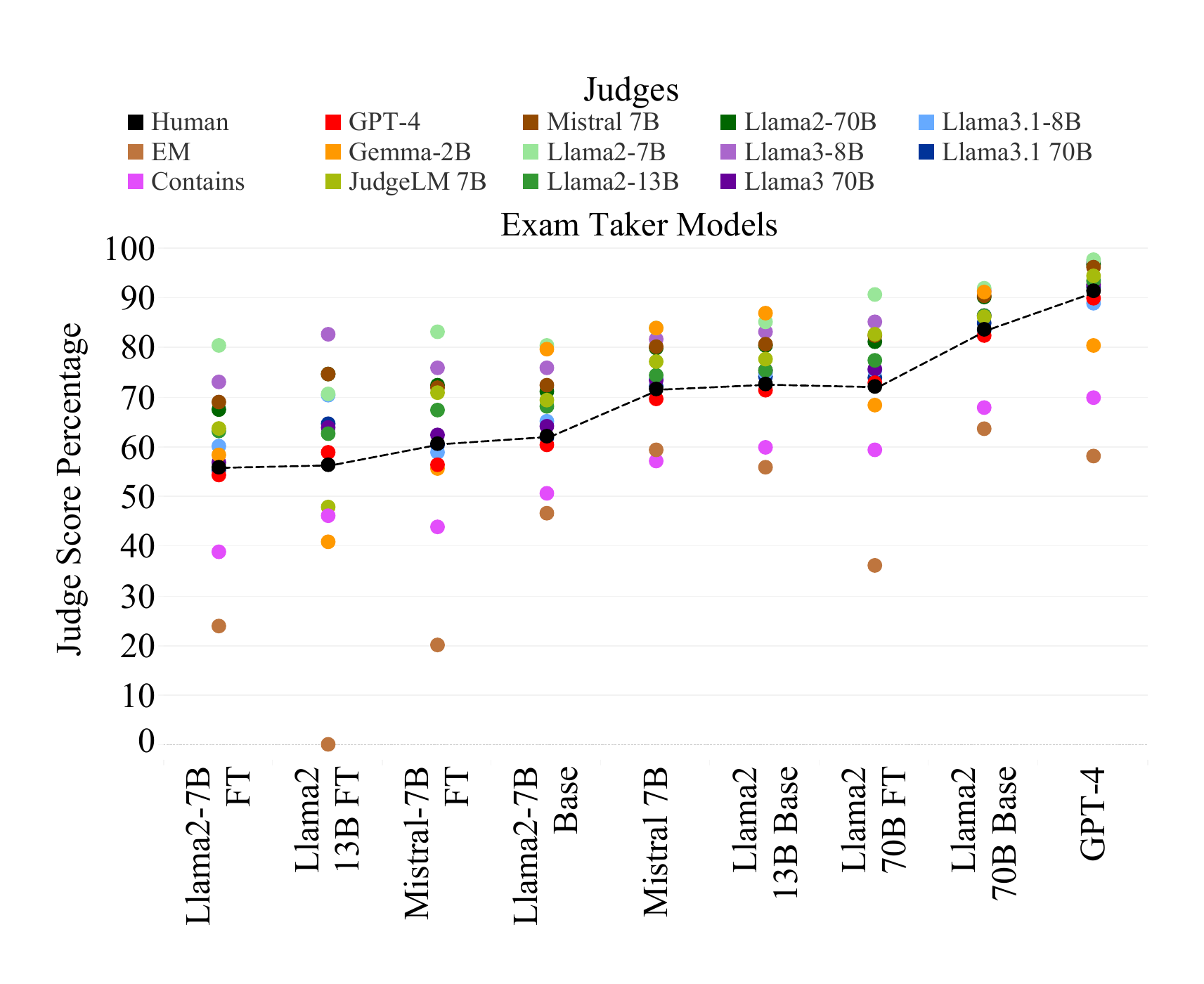}
        \vspace{-6mm}
        \caption{}
        \vspace{-2mm}
        \label{fig:llmalignment_a}
    \end{subfigure}
    \hfill
    \begin{subfigure}[b]{0.44\textwidth}
        \centering\includegraphics[width=\linewidth, height=6.5cm]{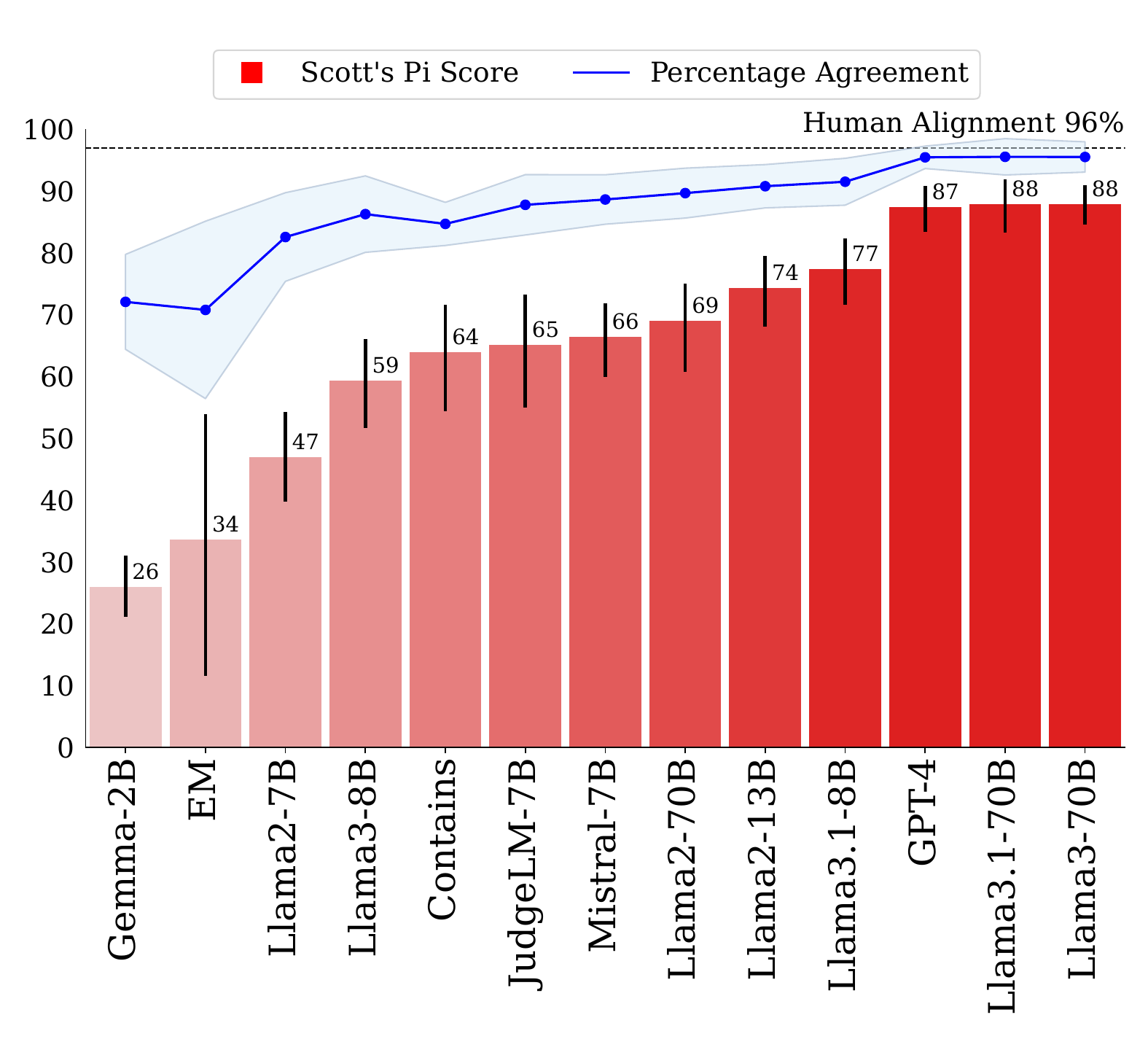}
        \vspace{-6mm}
        \caption{}
        \vspace{-2mm}
        \label{fig:llmalignment_b}
    \end{subfigure}
    \caption{\textbf{Average scores assigned by judge models and alignment with human judges.} (a) Scores assigned to all \evaluatormodels by the various \judgemodels. 
    (b) Average percent agreement (blue line) and Scott's $\pi$ scores (red bars) of \judgemodels with human judges (black line).
    Error bars annotate standard deviation across \evaluatormodels. 
    \judge{Llama3 70B}, \judge{Llama3.1 70B} and \judge{\gpt} have Scott's $\pi$ coefficient that are indicative of excellent alignment, but are still well below the human alignment score. 
    }
    \label{fig:llmalignment}
\end{figure*}

Over the last few years, large language models (LLMs) have demonstrated remarkable capabilities across various domains \citep[i.a.]{radford2019language, brown2020language, achiam2023gpt, meta2024llama3}.
%
As more and more new LLMs with different architectures and training methods continue to be released and their capabilities expand, accurately evaluating their performance and limitations becomes increasingly challenging \citep{zheng2024judging,ohmer2024form,benchekroun2023worldsense,madaan2024quantifying,li2023generative}.

LLM evaluation methods generally fall into one of two broad categories. Benchmarks such as MMLU \citep{mmlu}, TruthfulQA \citep{lin2021truthfulqa}, and GSM8K \citep{cobbe2021training} assess specific capabilities, while leaderboards such as Chatbot Arena \citep{chiang2024chatbot} and Open LLM Leaderboard \citep{open-llm-leaderboard} rank models based on human or automated pairwise comparisons. Both approaches face challenges in evaluating free-form text responses, as assessment can be as difficult as generation itself \citep[see e.g.][]{chang2023survey, bavaresco2024llmsinsteadhumanjudges}.


One approach to evaluating LLMs is using MCQ benchmarks like MMLU, which compare answer log-probabilities instead of assessing generated responses directly. However, this approach limits the range of measurable abilities and differs from how LLMs are used in practice. Lexical methods, such as exact match (EM) or n-gram overlap, are practical and cost-effective but prone to false negatives and often miss subtle semantic differences. These challenges are amplified for instruction-tuned chat models, which tend to produce more verbose responses \citep{saito2023verbosity, renze2024benefits}.

%
For these reasons, human evaluation remains the gold standard for evaluating LLM responses. 

\begin{table*}
    \centering
    \renewcommand{\arraystretch}{1.1} 
    \begin{tabular}{|>{\centering\arraybackslash}m{4.5cm}|>{\arraybackslash}m{9cm}|}
        \hline
        \textbf{\Evaluatormodels (base \& instruction-tuned)} & \eval{Llama-2 (7B, 13B, 70B)}, \eval{Mistral 7B}, \eval{\gpt} \\
        \hline
        \textbf{\Judgemodels (instruction-tuned)} & \judge{Llama-2 (7B, 13B, 70B)}, \judge{Llama-3 (8B, 70B)}, \judge{Llama-3.1 (8B, 70B)}, \judge{Gemma 2B}, \judge{Mistral 7B}, \judge{JudgeLM 7B}, \judge{\gpt} \\
        \hline
        \textbf{\Judgemodels (lexical)} & \judge{Exact Match (EM), Contains} \\
        \hline
    \end{tabular}
     \caption{\textbf{\Evaluatormodels and \judgemodels} We consider a wide variety of \evaluatormodels and \judgemodels; to get an in-depth overview of their abilities, we consider \evaluatormodels of various sizes \& types.}
    \label{tab:evaluation}
\end{table*}

Human evaluation is, however, expensive and often impractical, leading to the growing use of LLMs as \judgemodels \citep{lin2021truthfulqa,islam2023financebench,chiang2023can,liusie2024llm}. While promising alignment with humans has been noted \citep{sottana2023evaluation,zheng2024judging}, questions about this approach remain. This work examines LLMs as judges, contrasting them with humans and automated methods. Unlike prior studies, we focus on scenarios with high human alignment to separate task ambiguity from \judgemodel limitations. Using TriviaQA \citep{joshi2017triviaqa}, we evaluate how \textit{\judgemodels} of varying architectures and sizes assess \textit{\evaluatormodels}.

%
In this work, we study the properties of LLMs as judges, comparing them with humans and automated evaluation methods.
Contrary to prior work, we focus on a clean scenario in which human alignment is very high, 
allowing us to distinguish ambiguity and subjectivity in the task itself from potential issues with the \judgemodels.
Using the knowledge benchmark TriviaQA \citep{joshi2017triviaqa} as our playground, we investigate how \njudgesword different \textit{\judgemodels} with varying architectures and sizes judge \nexamtakersword different \textit{\evaluatormodels}.
    Our main findings are:
\begin{itemize}[leftmargin=4pt, topsep=1pt, itemsep=0.1em] 
    \item \textbf{Even in clean setups, only the best models have high alignment scores}. Among the \njudgesword \judgemodels, only \judge{\gpt}, \judge{Llama-3.1;70B}, and \judge{Llama-3;70B} achieved strong alignment with humans. However, even these fall short of the human alignment coefficient (\cref{fig:llmalignment}). 
\\
\item \textbf{\scottspi distinguishes judges better than percent alignment}. In terms of percent alignment, judges are rarely discriminable, while \scottspi provides a more informative signal. In some cases, high percent agreement can still give scores that differ 10-20 points from the human-assigned scores (\cref{fig:alignment_vs_delta}). 

\item \textbf{Also \scottspi is not all telling} While \judge{\gpt} and \judge{Llama-3} achieve excellent alignment scores, they can differ by up to 5 points from human scores. Moreover, in discriminating between \evaluatormodels, their performance is comparable to cheaper alternatives like \judge{Mistral 7B} and \judge{contains}, which have lower alignment scores but more consistent biases (\cref{fig:ranking_posneg}).


\end{itemize}

Through detailed analysis (\cref{sec:analysis}), we gain insights into judge performance. Improved alignment appear to be driven from higher recall rates and fewer false negatives. However, \judgemodels struggle with under-specified answers and exhibit leniency, reducing evaluation consistency. They are also sensitive to prompt length and quality. Surprisingly, even when asked to evaluate a verbatim match with a reference, \judgemodels sometimes fail.

Overall, our work highlights the strengths of the LLM-as-a-judge paradigm, while cautioning against overreliance on alignment metrics, even when they are high. Through error analysis, we identify common failure cases, contributing to a deeper understanding of this emerging evaluation paradigm. With this work, our objective is to improve understanding of the emerging mainstream paradigm for evaluating LLM.




\section {Related work}\label{sec:relatedwork}

Various recent studies have used or considered using LLMs as judges for tasks such as evaluating story generation \citep{chiang2023can}, retrieval-augmented generation \citep{es2023ragas},  visual QA \citep{maas2024improving}, code comprehension \citep{yuan2023evaluating}, multilingual evaluation \citep{hada2023large} and more general open-ended tasks \citep{zheng2024judging}.
\citet{Zhang2024LLMEval} and \citet{sottana2023evaluation} propose ways to standardise LLM evaluations and the role that \judgemodels might play in such solutions.
Several studies have demonstrated that state-of-the-art LLMs such as \gpt exhibit high alignment with human judgments \citep{sottana2023evaluation,zheng2024judging}, though others also illustrate that the paradigm is not yet without faults.
\citet{zeng2023evaluating} propose a benchmark for evaluating the performance of LLMs as judges, and other approaches have been proposed to improve LLM judges such that they are aligned well with humans \citep{shankar2024validates,zhu2023judgelm}.

Despite promising results in various settings, \judgemodels still suffer from known issues of current LLMs such as hallucinations and factual errors \citep{ye2023cognitive, turpin2023language} and difficulty in following complex instructions \citep{li2023instruction, he2024can}. 
Furthermore, various studies have reported challenges such as position bias \citep{pezeshkpour2023large,zheng2023large,wang2023large}, verbosity bias \citep{saito2023verbosity} in their preferences, confusing evaluation criteria \citep{hu2024llm}, or focusing more on the style and grammar compared to factuality \citep{wu2023style}.
Recently, \citet{liusie2024llm} have shown that LLMs perform better in comparative assessment compared to absolute scoring, which can be used for reliably measuring the relative performance of models \citep{liu2024aligning} and creating classifiers for pairwise grading 
\citep{llmasclassifier}.

We build on previous work to investigate the strengths and weaknesses of LLMs as judges. Unlike previous studies, we focus on comparing LLM outputs with reference answers rather than pairwise comparisons on open-ended tasks. With high human alignment in this setting, we gain a clearer view of LLM performance. Furthermore, we extend previous research by considering more LLMs, both as judges and as evaluated models.

\section{Methodology}\label{sec:methodology}


To evaluate the strengths and weaknesses of the LLM-as-a-judge paradigm, we focus on a comparatively controlled setup, in which \judgemodels assess answers of \evaluatormodels on the knowledge benchmark TriviaQA \citep{joshi2017triviaqa}.
With this methodological design, it is possible to focus on the abilities of the judges in isolation, without having to address human disagreement and error at the same time.
In this section, we elaborate the main aspects of our methodology.
\setlength{\parskip}{1pt}


\paragraph{Evaluation data}
%
As our testbed, we use the TriviaQA dataset \citep{joshi2017triviaqa}, consisting of 95K question-answer pairs sourced from 14 trivia and quiz league websites. 
Each question in the train and validation set is annotated with a list of short answers containing a minimal set of facts and evidence documents collected from Wikipedia and the Web.
For our experiments, we use the validation set of the \textit{unfiltered} partition of the benchmark, using the short answers as reference answers.
We use the training set for few-shot examples.

Since experiments require manual annotation of the \evaluatormodel responses, we use a random sample of $400$ questions from the dataset.
In \cref{app:downsamplingstddev}, we show with a bootstrapping test that this sample size has low variance for our main result.
Through experiments described in \cref{subsec:human_alignment}, we establish that humans have high agreement on  judgements of answers given to the questions in the benchmark.


\begin{figure*}
    \centering
    \begin{subfigure}[b]{0.415\textwidth}
        \centering
        \includegraphics[width=0.9\linewidth]{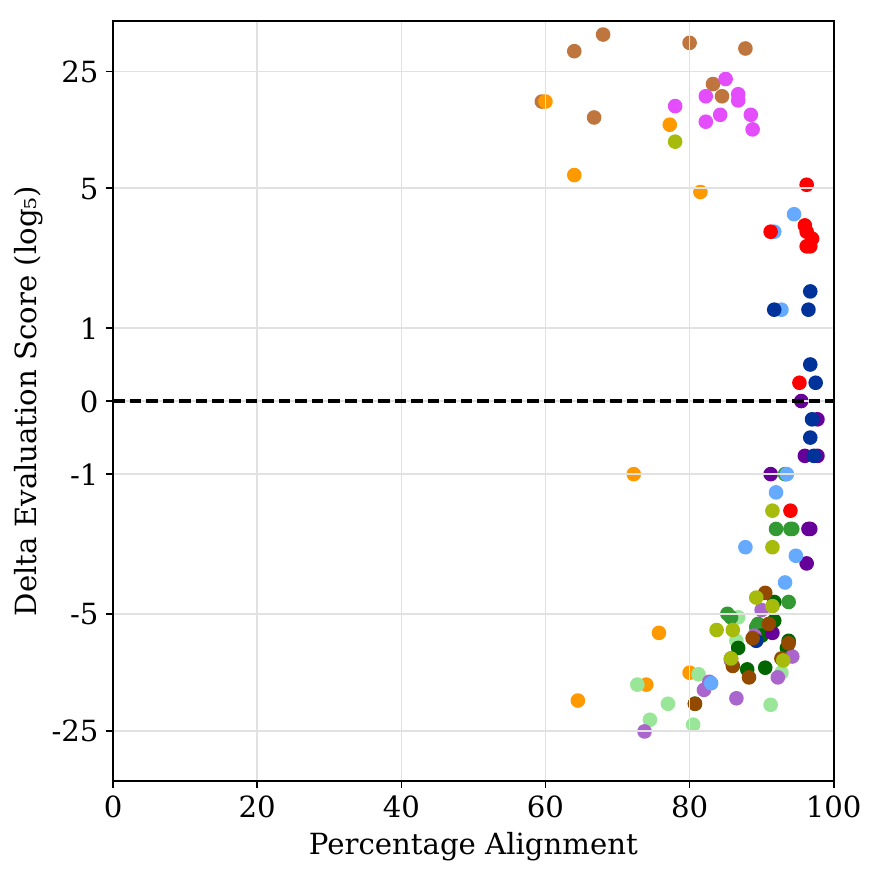}
    \end{subfigure}%
    \begin{subfigure}[b]{0.565\textwidth}
        \centering
        \includegraphics[width=0.9\linewidth]{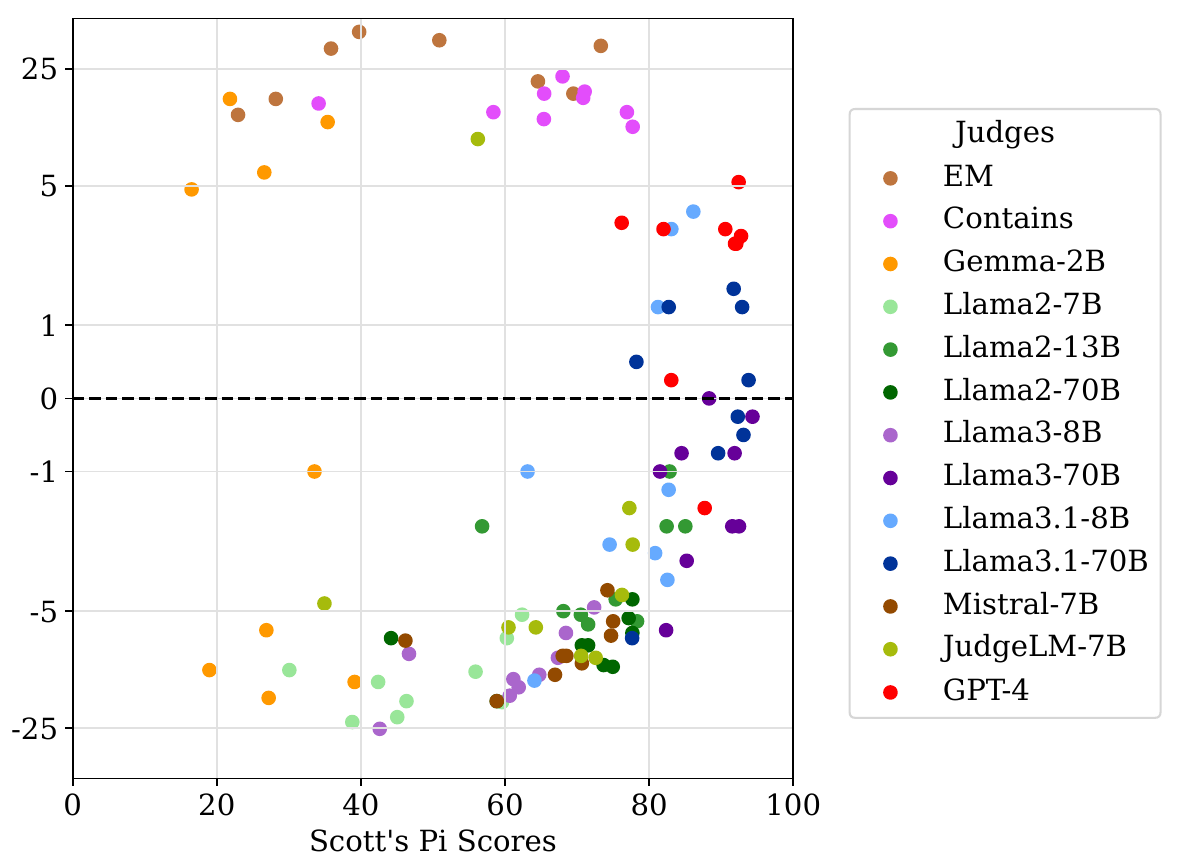}
    \end{subfigure}
    \caption{\textbf{Difference with human evaluation scores versus alignment metric.} 
    The delta evaluation score is the difference between the judge and the human score; y-axes are in log scale. 
    Percent alignment (left) shows a very skewwed distribution, making it difficult to distinguish models.
    \scottspi (left) provides a clearer difference between models, and is more indicative of deviation of the gold score.
    }
    \label{fig:alignment_vs_delta}
\end{figure*}

\paragraph{\Evaluatormodels} \label{subsec:evaluators}
%
To understand the strengths and weaknesses of different judges, we consider answers of pre-trained (base) and instruction-tuned (chat) `\evaluatormodels' across a wide variety of model sizes. 
In particular, we consider \eval{Llama-2} \citep{touvron2023llama} in 7B, 13B, and 70B parameter sizes for both base and chat versions, \eval{Mistral 7B} \citep{jiang2023mistral} base and chat versions, and \eval{\gpt}\footnote{Accessed via the OpenAI API between Mar 19th, 2024 and Sep 20, 2024.} \citep{achiam2023gpt} as the \evaluatormodels. 
The prompts for the \evaluatormodels contain five few-shot examples of (question, answer) pairs from the TriviaQA training set.
The prompts for the instruction-tuned models additionally include a command signaling the model to answer the given question in a succinct manner similar to the provided examples.
The prompts are provided in \cref{app:prompt-templates}.


\paragraph{\Judgemodels}
%
To get a comprehensive view of the strengths and weaknesses of \judgemodels across different model sizes and architectures, we use instruction-tuned versions of \judge{Llama-2} \citep{touvron2023llama} in 7B, 13B, and 70B sizes, \judge{Llama-3} \citep{meta2024llama3} in 8B and 70B sizes, \judge{Llama-3.1} \citep{dubey2024llama3herdmodels} in 8B and 70B sizes, \judge{Mistral} 7B \citep{jiang2023mistral}, \judge{\gpt} \citep{achiam2023gpt}, \judge{Gemma\;2B} \citep{gemma2024gemma}, and \judge{JudgeLM\;7B} \citep{zhu2023judgelm} as judges. To maintain parity with human and judge evaluation, judge prompts were built from human guidelines in \cref{app:human_annotation_guidelines}. The judges are instructed to respond with only a single word,  \texttt{``correct''} or \texttt{``incorrect''}.
An overview of all \evaluatormodels and \judgemodels is shown in \cref{tab:evaluation}.
For ease of reading, the \judge{\judgemodels} are depicted in a different font than the \eval{\evaluatormodels}.
\paragraph{Baselines} \label{subsec:baselines}
As baselines, we use two commonly used lexical evaluation techniques  -- exact match (\judge{EM}) and contains match (\judge{contains}).
For \judge{EM}, a response is considered correct if the response exactly matches one of the reference answers for the given question.
For \judge{contains}, an answer is considered correct if at least one of the reference answers is a sub-string of the response string.
Both EM and contains match are computed in a case-insensitive manner.

%
%

\paragraph{Alignment} \label{subsec:alignment}
We use two metrics to quantify alignment between judges: percent agreement and Scott's Pi coefficient \citep{scott1995scottspi}.%
\footnote{In an earlier version of this paper, we used Cohen's kappa \citep{cohen1960kappa} to measure alignment.
It has since come to our attention that -- despite it's widespread use -- this metric has some well-documented theoretical issues \citep[e.g.][]{pontius2011death,chicco2021matthews}.
For the interested reader, we elaborate on these issues in \cref{app:cohenslimitation}.
}
Percent agreement expresses a simple percentage of the samples on which two annotators agree. 
Scott's Pi, denoted as \scottspi, is an alignment metric that corrects for chance agreement between two annotators and is considered to provide a more robust measure of alignment.
Details about the computation of both metrics are given in \cref{app:metrics}.

\begin{figure*}[t]
    \centering
    \begin{subfigure}[b]{0.46\textwidth}
        \centering
        \includegraphics[width=\linewidth]{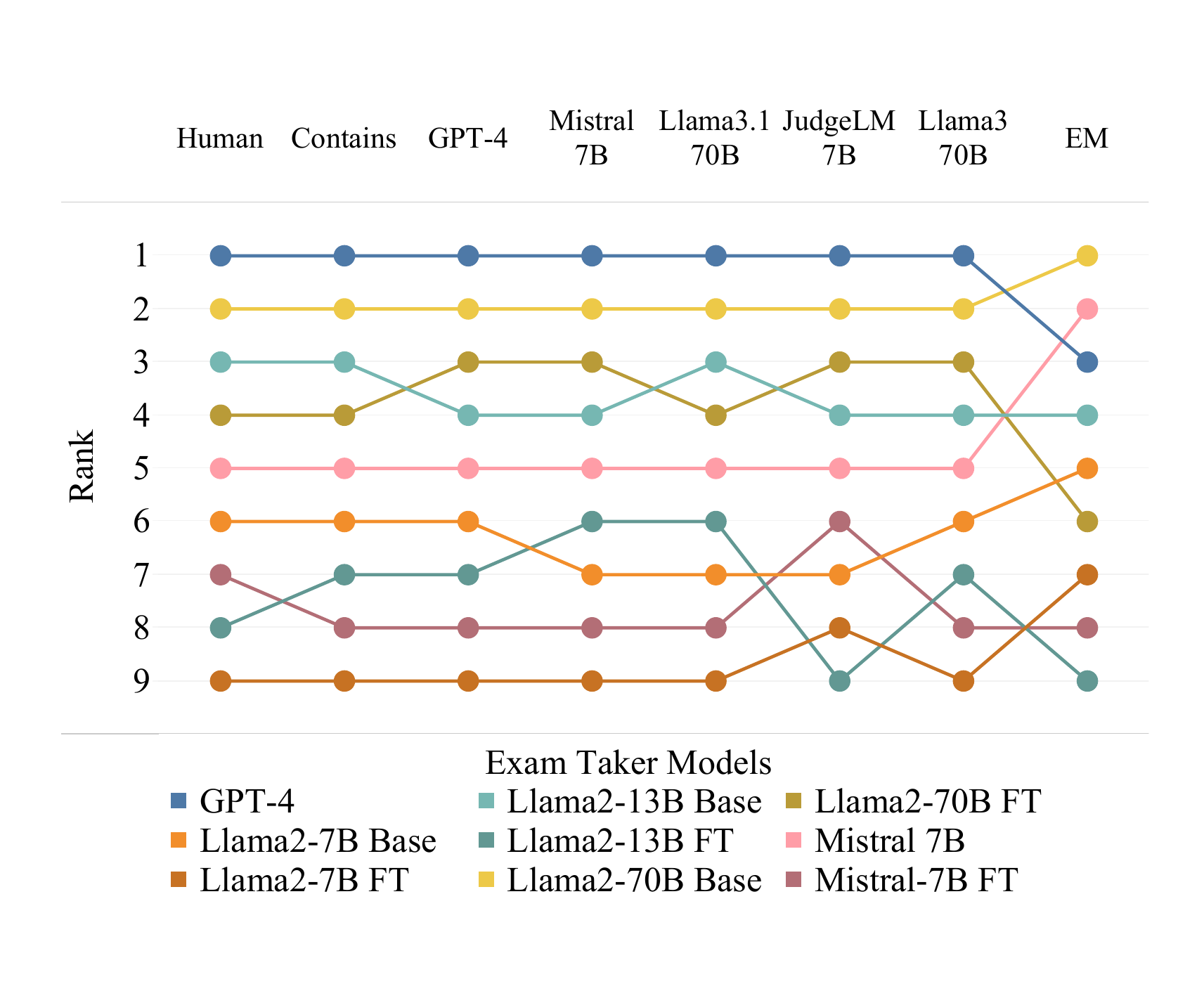}
        \vspace{-8mm}
        \caption{}
        \vspace{-2mm}
        \label{fig:rankcorrelation}
    \end{subfigure}
    \hfill
    \begin{subfigure}[b]{0.53\textwidth}
        \centering
        \includegraphics[width=\linewidth, height=5.5cm]{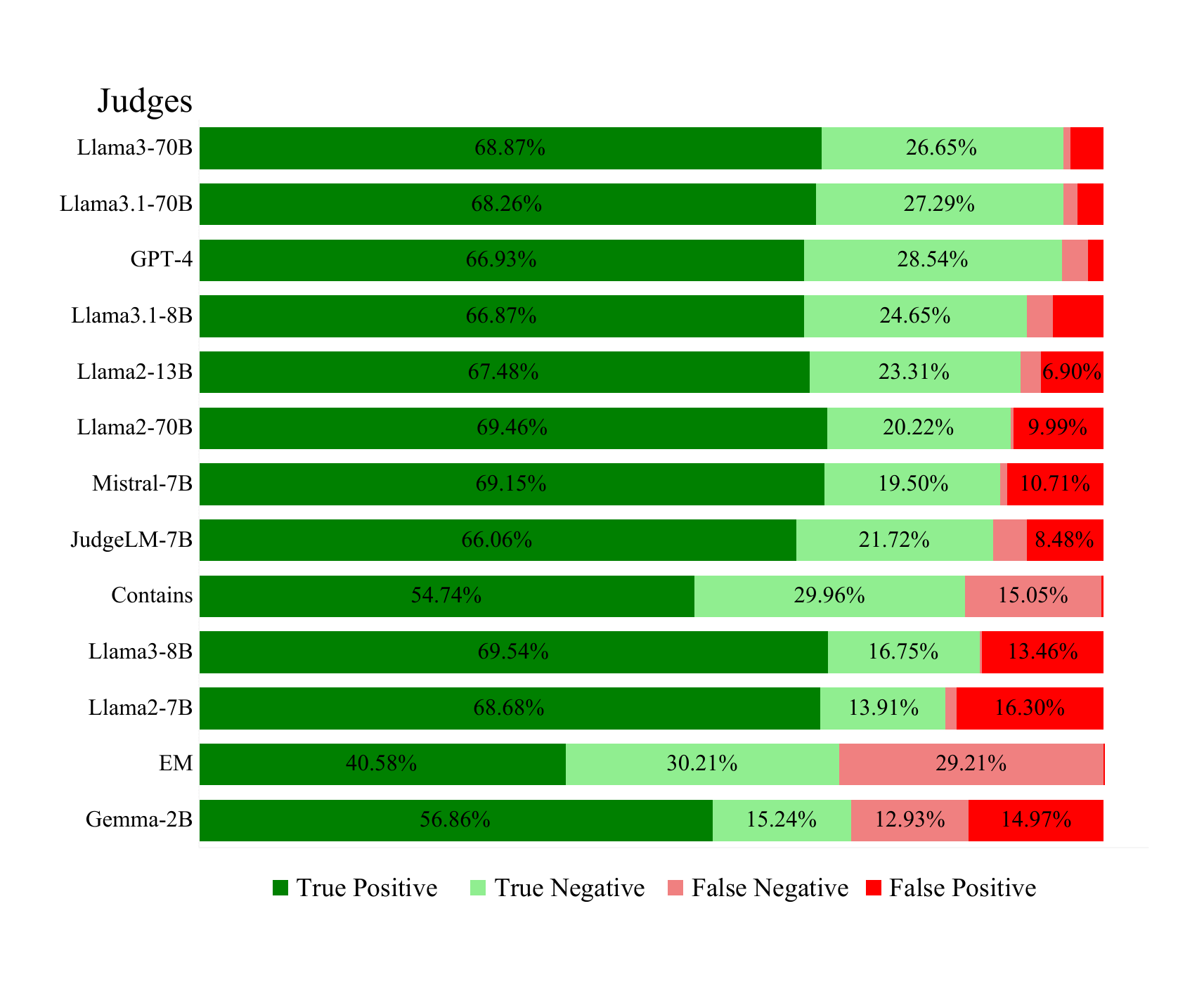}
        \vspace{-8mm}
        \caption{}
        \vspace{-2mm}
        \label{fig:confusionmatrix}
    \end{subfigure}
    \caption{\textbf{Judge rankings and true/false positives and negatives.} 
    (a) Assigned \evaluatormodel rankings assigned by highly human aligned judges.
    \judge{Contains} stays closely to human-assigned rankings, as well as \judge{\gpt} and \judge{Mistral 7B}. 
    (b) False positives and negatives across different \judgemodels, in descending order of human alignment. 
    Both false negatives and false positives increase as human alignment decreases, but well-aligned models tend to produce more false positives than false negatives.
        }\label{fig:ranking_posneg}
\end{figure*}

\paragraph{Human judgements} \label{subsec:human_alignment}
As a ground-truth assessment, we obtain human annotations for each \evaluatormodel answer.
The inter-human alignment is calculated between three human judges using the answers to 1200 randomly sampled questions answers; the human guidelines can be found in \cref{app:human_annotation_guidelines}. 
We then determine collective ``Human Judgment'' through a majority vote.
%


The average alignment between human evaluators and the majority vote yielded a \scottspi of $96.2\pm1.07$,%
\footnote{The coefficient is scaled by $100$ for easier comparison with percentage alignment.}
while the average percentage agreement was $98.52\%\pm0.42\%$,  exceeding the alignment previously reported in comparable studies \citep{zeng2024evaluatinglargelanguagemodels}.

The details of this experiment are mentioned in \cref{app:limitations}.
Given this near-perfect alignment score, we consider only one human evaluator per sample for the rest of our experiments, to reduce the overall cost of human annotations. 
The set of questions for which we obtain human annotations is identical for each \evaluatormodel.
%

%

\section{Results} \label{sec:results}

In this section we discuss our main results, primarily focusing on the relationship between evaluations by various \judgemodels and human evaluations (\cref{sec:results:exploringhumanjudgellmalignment}), and how that impacts their usability (\cref{sec:results:exploringsystematicpatterns}).
To do so, we evaluate their alignment with human judgment and assess how differently they rank the \nexamtakersword \evaluatormodels compared to humans.
In Section \ref{sec:analysis}, we further analyse their precision and recall to further investigate the types of errors that can be made by various \judgemodels. Details about compute requirements and others costs for experiments are given in \cref{app:experiment-costs}.

\subsection{Alignment between \judgemodels and humans}
\label{sec:results:exploringhumanjudgellmalignment}

We start by computing \scottspi scores and percent agreement between the evaluations of each \judgemodel and the human annotators. 
We show the result in \cref{fig:llmalignment}.
We observe that percent alignment is high for virtually all models, with the exception of \judge{Gemma\;2B} and \judge{EM}.
\scottspi, on the other hand, has low values for most models, though its value is in the high 80s for \judge{Llama-3\;70B}, \judge{Llama-3.1\;70B}  and \judge{\gpt}. 
Nevertheless, there still is a significant disparity between human judgment and \judgemodels: the best scoring judge, \judge{Llama-3\;70B}, is 8 points behind human judgment. 
Notably, \judge{EM} has the most variance in alignment, while \judge{Gemma\;2B} has the lowest alignment amongst all judges.


In most cases, we observe that \scottspi and percent agreement are following the same trend, with the exception of the values for \judge{Gemma\;2B} and \judge{EM}.
\judge{Gemma\;2B} shows higher percent agreement compared to \judge{EM}, yet it yields the lowest \scottspi score within the ensemble.
For the percent agreement of judge models, we note a 26-point difference between human judgment and EM, while \scottspi exhibits a more substantial 64-point gap. 
This is also visible in the general decline of alignment scores: while \judge{Llama-3\;8B} has a \scottspi score of only 59, its percent agreement is still well above 80\%.
Overall, \scottspi appears to be better able of discriminating various judge models, showing more divergence across the tested judges.

To understand how indicative the two alignment metrics are of the expected accuracy of the overall judgement of the models, we plot, for each \judgemodel and \evaluatormodel, the difference between the score assigned by the judge and the score assigned by a human.
In the figure, we can see that for \scottspi values higher than 80, the evaluation scores are comparatively close to the human evaluation scores, with a difference of up to 5 points in their assigned scores (complete results table provided in \cref{app:all_scores}). 
For percent alignment, on the other hand, even judges that have more than 90\% may still differ more than 10 points in their assigned score.
%
Interestingly, 
the deviation from human-judgements for a single judge model can be quite different depending on the \evaluatormodel.
In \cref{fig:llmalignment_a}, \judge{Gemma\;2B}, for instance, sometimes assigns higher scores than humans, and sometimes much lower. 
In the next section, we further explore this particular pattern.

\subsection{Exploring consistent patterns in \judgemodels} \label{sec:results:exploringsystematicpatterns}


In the previous section, we saw that none of the \judgemodels were as aligned with humans as humans were with each other. As shown in \cref{fig:alignment_vs_delta}, even the best-aligned \judgemodels can differ by up to 5 points from human-assigned scores. While this limits their ability to perfectly estimate \evaluatormodel capabilities, \judgemodels can still provide valuable insights to \textit{differentiate} between \evaluatormodels. For example, judges with consistent biases may not assign identical scores but could rank models similarly, akin to a very strict teacher.

To assess this, we compare the rankings given by each \judgemodel to the nine \evaluatormodels, computing Spearman's rank correlation coefficients $\rho$ \citep{spearman1904spearman} with the human ranking. The rankings are shown in \cref{fig:rankcorrelation}, with $\rho$ and $\sigma$ values in \cref{app:correlationcoefftable}. Most \judgemodels have rank correlations above 0.7, indicating they struggle to distinguish poorer models but do well with better ones. Notably, models like \judge{contains} and \judge{Mistral 7B}, which have divergent scores from humans, show high rank correlation ($\rho$ of 0.99 and 0.98, respectively), performing similarly to \judge{\gpt} and outperforming the better \judge{Llama} models -- though with lower significance values -- indicating that identifying which models are better should not be equated to assigning them the correct score.

\section {Analysis}\label{sec:analysis}

\begin{figure*}[t]
    \centering
    \begin{subfigure}[b]{0.48\textwidth}
        \centering
        \includegraphics[width=\linewidth]{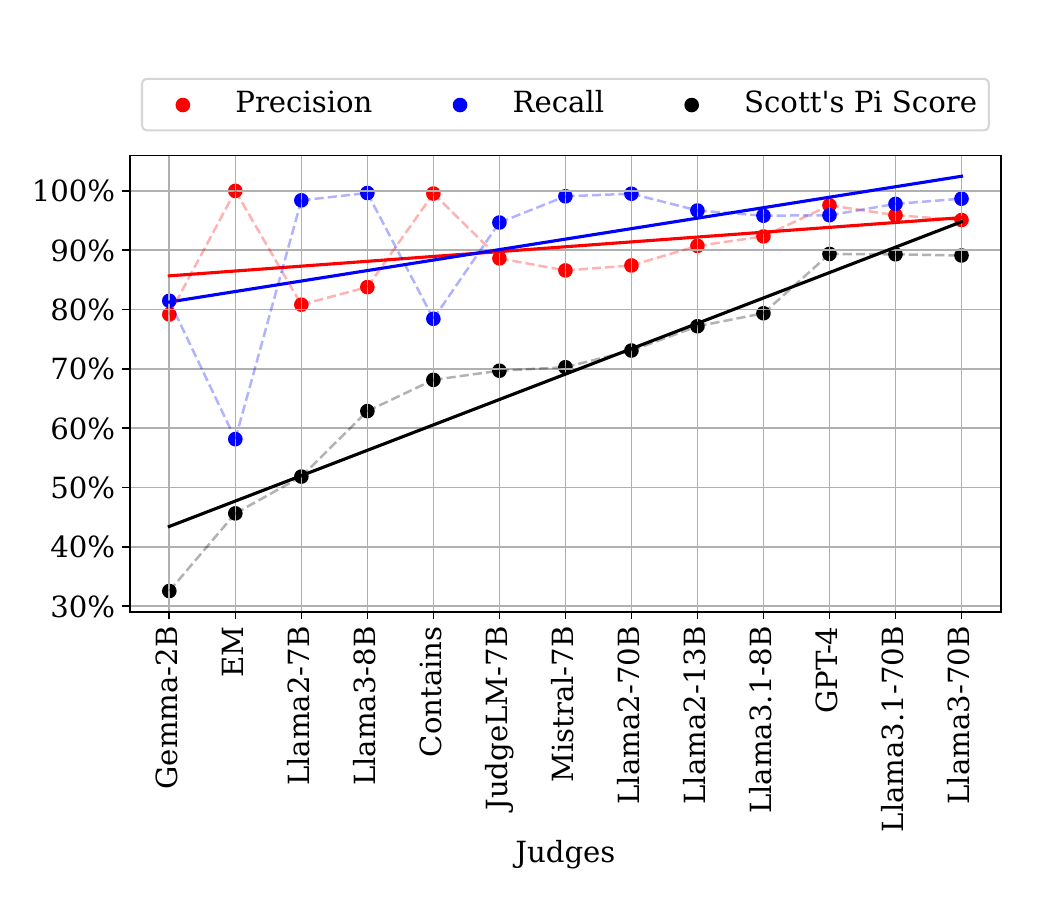}
        \vspace{-6mm}
        \caption{}
        \vspace{-2mm}
        \label{fig:precisionrecall}
    \end{subfigure}
    \hfill
    \centering
    \begin{subfigure}[b]{0.51\textwidth}
        \centering
        \includegraphics[width=\linewidth]{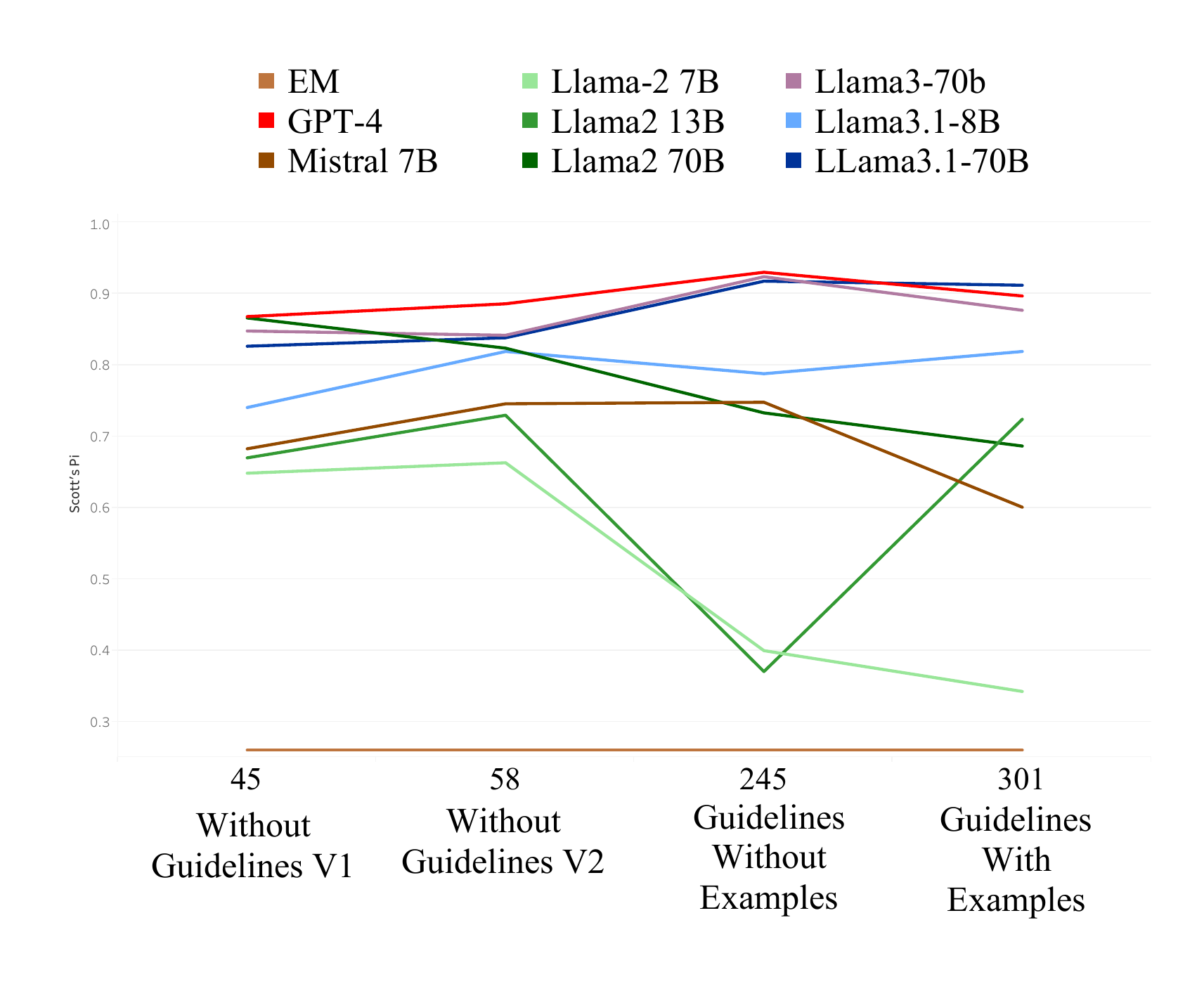}
        \vspace{-6mm}
        \caption{}
        \vspace{-2mm}
        \label{fig:TooMuchInfo}
    \end{subfigure}
    
    \caption{\textbf{Precision, recall and prompt sensitivity.} (a) Recall and precision improve with increasing human alignment (\( R^2 \) = 0.31 and \( R^2 \) = 0.21, respectively). 
    (b) \scottspi scores for judges across different instructions. 
    }
\end{figure*}

\begin{table*}[t]
\resizebox{\textwidth}{!}{
    \begin{tabular}{lllllll}
        \textbf{Error code} & \textbf{Explanation} & \textbf{Example} & \textbf{Proportion} & \textbf{GPT-4 recall} & \textbf{Llama-3 70B recall}\\ 
        \toprule\toprule
        \textbf{Incorrect entity} & \specialcell{Response refers to a wrong entity} & \specialcell{\texttt{Henry VII, James I, Edward VI,} \\ \texttt{Mary I and Elizabeth I}} & 86.9\% & \textbf{98.3\%} & \textbf{96.6\%}\\
        \hline
        \textbf{Under-specified} & \specialcell{Response contains only part \\ of the answer} & \specialcell{\texttt{Henry VII, Henry VIII, Edward,} \\ \texttt{Mary, and Elizabeth}} & 37.3\% & 33.9\% & 23.3\% \\
        \hline
        \textbf{Too few entities} & \specialcell{Response contains too few entities} & \specialcell{\texttt{Henry VII, Edward VI,} \\ \texttt{Mary I and James I}} & 2.47\% & \textbf{80.0\%} & 60.0\%\\
        \hline
        \textbf{Too many entities} & \specialcell{Response contains too many entities} & \specialcell{\texttt{Henry VII, Henry VIII, Edward VI,} \\ \texttt{Mary I, James I, and Elizabeth I}} & 2.7\% & \textbf{90.1\%} & \textbf{90.1\%} \\
        \hline
        \textbf{Other} & \specialcell{Response is incorrect but cannot \\ be put into any of the above buckets} & \specialcell{\texttt{I'm sorry but I do not know the} \\ \texttt{answer to that question}} & 1.23\% & 20.0\% & 40.0\% \\
        \bottomrule
    \end{tabular}
    }
     \caption{\textbf{Error analysis for \judge{GPT-4} and \judge{Llama-3 70B} judges.} 
     The example question is \textit{``Excluding Lady Jane Grey, who were the five monarchs of the House of Tudor?''}, the correct answer \textit{``Henry VII, Henry VIII, Edward VI, Mary I and Elizabeth I''} (in any order).}
     \label{table:error_codes}
\end{table*}

To better understand the \judgemodels, we conduct multiple case studies aimed at identifying common errors and vulnerabilities in the judges we investigate.
Specifically, we study their precision and recall and error types (\cref{sec:analysis:subsec:precision_recall}), their sensitivity to the instruction prompt prompt (\cref{sec:analysis:subsec:instructions}), how they respond to controlled resposes of specific types (\cref{sec:analysis:subsec:judge-ability}), and the extent to which they have a \textit{leniency bias} (\cref{sec:leniency-bias}).

\subsection{Better aligned models: Precision and recall gains with error spotlights}
\label{sec:analysis:subsec:precision_recall}

We first examine the precision and recall of the \judgemodels. As shown in \cref{fig:precisionrecall}, both metrics increase moderately with alignment. \cref{fig:confusionmatrix} reveals a similar trend, with a clearer distribution of false positives and negatives. True positives remain consistent across varying judge quality, whereas true negatives exhibit a slight decline as judge quality decreases. Notably, a reduction in judge quality leads to an increase in false positives.


Next, we analyze the errors made by \judgemodels by manually annotating 900 outputs from \eval{Llama-7B Base}, focusing on top performers \judge{\gpt} and \judge{Llama-3;70B}. We categorize error types and determine how often they are correctly judged as incorrect. The results in \cref{table:error_codes} show that both \judge{\gpt} and \judge{Llama-3;70B} excel at identifying answers referring to incorrect entities or containing too many entities. Under-specified and incorrect answers are more challenging, with \judge{\gpt} performing better on answers with fewer entities than \judge{Llama-3;70B}.

\begin{figure*}[t]
    \centering
    \includegraphics[width=0.825\textwidth]{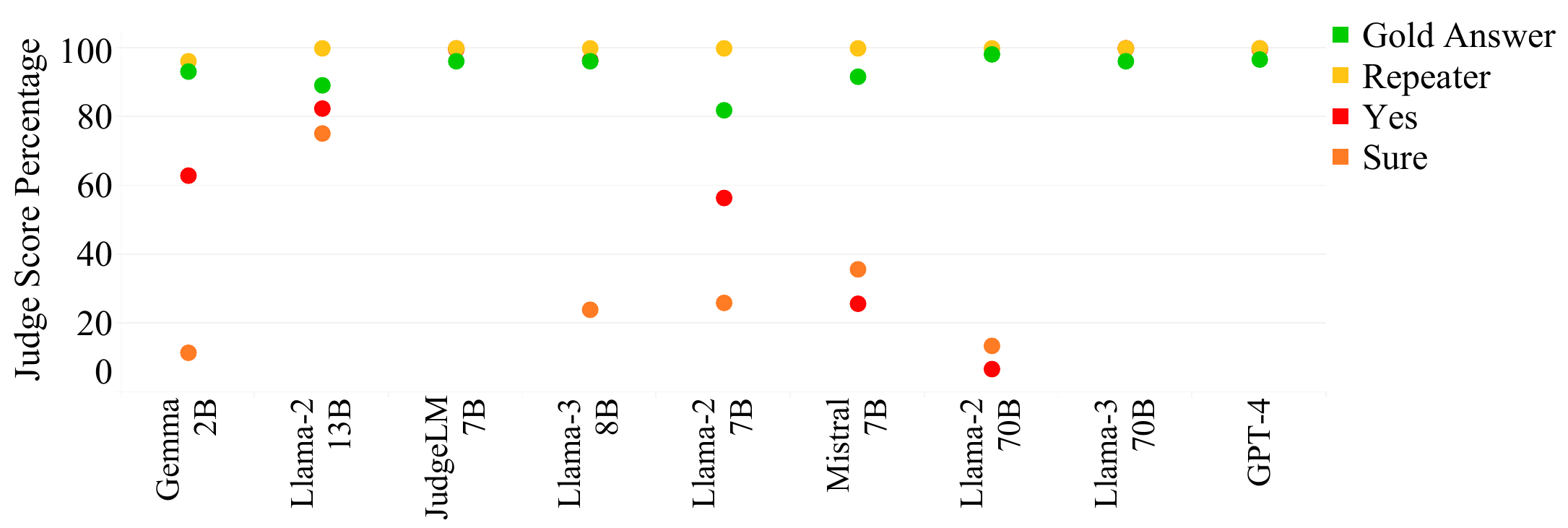}
    \caption{\textbf{Judge responses to dummy answers.} 
    We investigate how \judgemodels respond to dummy answers.
    \judgemodels remain robust when \evaluatormodels produce responses identical to the prompt (`repeater'), but are less robust when the responses are "Yes" and "Sure". Even when the answer matches one of the reference answers verbatim (`Gold answer'), judges do not always arrive at the correct judgement.
    }
    \label{fig:judge_dummy}
\end{figure*}

\subsection{\Judgemodel sensitivity to prompt length and specificity}
\label{sec:analysis:subsec:instructions}

Next, we investigate how prompt length and specificity affect \judgemodels' inferences to determine whether their performance is influenced by \textit{specificity} of the prompt. We use four prompt versions with varying length and specificity.

The first two prompts (\texttt{Without;guidelines;V1/V2}, 45 and 58 tokens) ask for an evaluation without further details. The longer prompts (\texttt{Guidelines;without;examples} and \texttt{Guidelines;with;examples}, 245 and 301 tokens) provide more elaborate guidance and examples. All prompts are listed in \cref{app:TMI}.

\cref{fig:TooMuchInfo} shows that \judge{\gpt}, \judge{Llama-3;70B}, and \judge{Llama-3.1;70B} exhibit low variance in human agreement as prompt length and specificity increases. Top performers show high alignment with humans even with minimal instructions, while they slightly improve with more detailed prompts. In contrast, other models lose alignment with increased instructions, likely due to difficulty processing complex instructions.

In a follow-up experiment, we investigate the impact of reference order (see \cref{app:ref-bias-exp}). \cref{app:ReferenceBiasExample1} and \cref{app:RefernceBiasExample2} shows that larger models maintain consistent judgments regardless of reference order, while smaller models, except \judge{Mistral;7B}, are more sensitive to it.

\subsection{Evaluating controlled responses}
\label{sec:analysis:subsec:judge-ability}

We conduct simple tests on the \judgemodels by having them evaluate dummy benchmark responses. In the first test, the answer is a verbatim reference from the dataset (always correct). In the next three tests, the answers are incorrect. For the second and third tests, the dummy \evaluatormodel responds with \texttt{``Yes''}, and \texttt{``Sure''} respectively. In the fourth test, the evaluated answer is a repetition of the question.


%

In \cref{fig:judge_dummy}, we observe that while some \judgemodels correctly identify and mark answers as correct (first test) or incorrect (next three tests), others, like \judge{Llama-2;70B}, incorrect evaluate many dummy answers, despite showing high human alignment on benchmark evaluations (see \cref{fig:llmalignment_b}). We hypothesize that when the answers are plausible but incorrect, judges can correctly identify them as wrong by comparing them with the reference. However, when the answer is unrelated (e.g., \texttt{``Yes''}, and \texttt{``Sure''}), \judgemodels may mistakenly mark them as correct, though further research is needed to clarify this behavior.

\subsection{Leniency bias in \judgemodels}
\label{sec:leniency-bias}

    

Lastly, to get a general sense of the inherent biases or misalignment in the evaluation criteria that might be present in the judge models, we estimate if they have a positive or negative bias in their judgment.
To do so, we assume that a judge assigns the correct judgment (i.e.\ same evaluation as the ground truth) with a probability of $P_c$ and assigns the rest of the samples to be \texttt{``correct''} with a probability $P_+$, which we call their \textit{leniency bias}.
We estimate the values of $P_c$ and $P_+$ from the benchmark results\footnote{
The theoretical derivation of the expressions for $P_c$ and $P_+$, as well as the empirical validation for their estimated values can be found in \cref{app:leniency-bias}.}
and show them in \cref{tab:p-vals-full}. 
We observe that $P_+$ for most models is significantly higher than $0.5$ (\cref{fig:k-p-corr}), indicating a tendency of the \judgemodels to evaluate responses as \texttt{``correct''} when their evaluation criteria are not completely aligned with the provided instructions. 

\section {Conclusion}\label{sec:conclusion}

In this work, we conduct an extensive study of LLMs as judges, comparing them to human judges and automated evaluation methods. By focusing on a clean evaluation scenario with high inter-human agreement, we identify potential issues with the LLM-as-a-judge paradigm, separate from task ambiguity.

We find that smaller, cost-efficient models, like \judge{Mistral;7B}, are less effective than larger models such as \judge{\gpt}, \judge{Llama-3.1;70B}, and \judge{Llama-3;70B}, which are better aligned but still fall short of human alignment. Even with high alignment, their scores can differ by up to 5 points from human scores, highlighting the need for caution when using judges in more complex scenarios. We also note that the commonly used metric of \textit{percent aligned} fails to differentiate between judges effectively. We suggest future work adopt the more robust \scottspi metric for better distinction.

Next, we note that high alignment scores are not always necessary to \textit{discriminate} between models. While \judge{\gpt} and \judge{Llama-3} have excellent alignment scores, simpler and more cost-efficient models, like \judge{contains}, perform similarly in ranking \evaluatormodels, despite lower alignment scores and score deviations. For studies focused on ranking models rather than estimating exact scores, these approaches can be as suitable as more expensive ones.

Lastly, we run experiments to assess judge models' sensitivity to prompts, precision, recall, error types, leniency, and vulnerability to dummy answers. We find that smaller models are more likely to judge positively when in doubt, that lower-alignment models lack precision, and that better models are more robust across different prompts but harder to "steer." Some \judgemodels are easily fooled by dummy answers like \texttt{''Yes''} and \texttt{''Sure''} and are better at detecting completely incorrect answers than partially incorrect ones.

Overall, this work contributes to LLM evaluation by assessing judges in a clearly defined framework. Our results highlight the potential of LLMs as judges but caution against blindly trusting their judgments, even when aligned with humans. We recommend computing both percent agreement and \scottspi, paired with qualitative analysis, to avoid bias. We discuss limitations in \cref{app:limitations} and plan to expand our work to more complex scenarios in the future.

\bibliography{bibliography}

\appendix
\newpage
\appendix
\renewcommand{\thesection}{\Alph{section}}

\section{Limitations}\label{app:limitations}

In our work, we have evaluated how 11 different LLMs fare as judges in a scenario in which judgements should be relatively straight-forward, and human alignment is high.
As any study, our work has several limitations as well as directions that we did not explore but would have been interesting too.
In this section, we discuss both.

\paragraph{Simplicity of the task}
As mentioned in the introduction of our work, the scenario in which judges are used are typically much more complicated than the scenario that we focussed on.
Specifically, judges are most often deployed in preference rankings (where two model responses are compared) or to judge complex answers that are difficult to automatically parse.
In such tasks, human agreement is often low, making it challenging to judge the judges themselves.
In our work, we have deliberately chosen for a simple task, in which human alignment is high.
The main premise is, that if a judge does not perform well in this simple setup, caution is suggested also in more complex setups -- if someone cannot do multiplication, why would they be able to solve ordinary differential equations.
Given the poor understanding of which abilities of LLMs generalise in what dimensions, however, more studies are needed to understand how our results generalise to various other scenarios.

\paragraph{Human alignment}
In an earlier version of this paper, due to the high cost of human annotations, we opted to select a single model for human annotation as we iteratively modified the exam taker prompt, few-shot examples, and guidelines. 
We selected the \eval{Llama2 7B} for this purpose with a random sample of 600 questions.
As this is only a single model, it is possible that our human alignment scores are biased because of that.
After, we have therefore extended our results with another 600 human-annotated examples from \eval{Llama3.1 70B}.

For \eval{Llama2 7B} The average alignment among human evaluators had a \scottspi of $96.36\pm1.46$,%
and the average percent agreement was $98.33\%\pm0.76\%$.
For \eval{Llama3.1 70B}, we noted that the average alignment among human evaluators had \scottspi of $95.78\pm0.30$,\% and the average percent agreement was $98.72\%\pm0.10\%$.
Given the similarity of these two numbers, we believe that these 1200 samples provide an adequate estimate.
In the paper, we take the average.

\paragraph{Size of the judged samples}
As each of the \nexamtakersword \evaluatormodels requires human annotations for each sample, we restricted our analysis to 400 samples in total.
This sample size also allowed us to conduct manual annotations and error analysis within 75 human hours/200 GPU hours (see \cref{app:experiment-costs}) and give reliable confidence intervals while also providing the flexibility to compare a range of models. 
We were not able to increase the size due to the cost, but a statistical analysis (details provided in \cref{app:downsamplingstddev}) illustrated that the variance because of this sample size was very low.

\paragraph{Selection of judges}
With our selection of judges, we have stuck to autoregressive judges that can be used off-the-shelve, as well as one LLM specifically trained to judge.
They are -- at the moment of writing -- the ones that are most commonly used as LLM-judges, and we have tried to be comprehensive across size and family.
Nevertheless, we acknowledge that there are other judges that we could have considered as well.
As including more judges in -- compared to including more \evaluatormodels -- relatively straightforward because it requires only computational power, no manual annotation, we hope that others may evaluate their newly proposed judges using our setup as well.

\paragraph{Future work}
All in all, these differences underline how finicky using LLMs as judges can be, and with that confirm the overall conclusions of our study that much more work is needed to better understand the strengths and limitations of judge models across a wide range of scenarios and model accuracies.
We consider assessing the strengths across multiple different samples and tasks, which would require many more human annotations, outside the scope of this paper and leave such experimentation for future work.

\section{A brief explanation of the theoretical issues with Cohen's kappa}\label{app:cohenslimitation}

Cohen's Kappa Coefficient \citep{cohen1960kappa} is a statistic to measure inter-rater agreement for categorical responses.
Cohen's Kappa coefficient measures this agreement by computing the observed (percent) agreement between raters ($p_o$) and comparing it with the hypothetical probability of chance agreement ($p_e$), which is taken as a baseline, as follows:
\begin{equation}
\kappa \equiv \frac{p_o - p_e}{1 - p_e}
\end{equation}

In this equation, the chance agremeent $p_o$ constitutes the hypothetical probability that observed agreement occurred by chance, given the observed distributions of the considered raters, under the assumption that the probabilities the raters assign to the observed labels are independent.
Specifically, it is defined as:

\begin{equation*}
\begin{aligned}
p_e &= \sum_k \widehat{p_{k12}} =^{ind} \sum_k \widehat{p_{k1}} \widehat{p_{k2}} \\
    &= \sum_k \frac{n_{k1}}{N} \cdot \frac{n_{k2}}{N}
     = \frac{1}{N^2} \sum_k n_{k1} n_{k2}
\end{aligned}
\end{equation*}

where $\widehat{p_{k12}}$ is the estimated probability that rater 1 and rater 2 will classify the same item as $k$, rewritten to $\widehat{p_{k1}}\widehat{p_{k2}}$ under the assumption that $p_{k1}$ and $p_{k2}$ are independent.
The crux of the issue with this method of computation, is that $\widehat{p_{k1}}$ and $\widehat{p_{k2}}$ are estimated independently from the data.
As such, the chance agreement adjusts for the observed average differences between raters, which is in fact part of what we intend to measure.

To address this issue, Scott's Pi \citep{scott1995scottspi} instead defines the chance baseline under the assumption that the raters have the same distribution, which is estimated considering the joint distribution of rater 1 and rater 2, rather than considering them separately.
It defines $p_e$ as:

\begin{equation}
p_e = \sum_k \widehat{p_k^2} = \sum_k \sum_k (\frac{n_{k1} + n_{k2}}{2N})^2
\end{equation}

As such, contrary to Cohen's Kappa, it captures differences surpassing the chance agreement if rater 1 and rater 2 were in fact equivalent.
In other words, we compare against a baseline in which raters would be equivalent, and we measure how much they deviate from that.

Note that if the empirical distributions of rater 1 and rater 2 are the same, so will the values of Scott's Pi and Cohen's Kappa be.
This also implies that for larger observed (percent) alignment values, the values for Cohen's Kappa and Scott's Pi will be closer.

\section{Model and dataset details}\label{app:asset-details}

In \cref{tab:assets}, we show the different models and datasets used in our experiments, along with version and license details.

\begin{table*}[h]
    \centering
    \begin{tabular}{lll}
        \toprule
        Asset & Version & License \\
        \midrule
        TriviaQA & \href{https://huggingface.co/datasets/mandarjoshi/trivia_qa}{\texttt{mandarjoshi/trivia\_qa}} & apache-2.0 \\
        Llama-2 7B Base & \href{https://huggingface.co/meta-llama/Llama-2-7b-hf}{\texttt{meta-llama/Llama-2-7b-hf}} & llama2 \\
        Llama-2 7B Chat & \href{https://huggingface.co/meta-llama/Llama-2-7b-chat-hf}{\texttt{meta-llama/Llama-2-7b-chat-hf}} & llama2 \\
        Llama-2 13B Base & \href{https://huggingface.co/meta-llama/Llama-2-13b-hf}{\texttt{meta-llama/Llama-2-13b-hf}} & llama2 \\
        Llama-2 13B Chat & \href{https://huggingface.co/meta-llama/Llama-2-13b-chat-hf}{\texttt{meta-llama/Llama-2-13b-chat-hf}} & llama2 \\
        Llama-2 70B Base & \href{https://huggingface.co/meta-llama/Llama-2-70b-hf}{\texttt{meta-llama/Llama-2-70b-hf}} & llama2 \\
        Llama-2 70B Chat & \href{https://huggingface.co/meta-llama/Llama-2-70b-chat-hf}{\texttt{meta-llama/Llama-2-70b-chat-hf}} & llama2 \\
        Mistral 7B Base & \href{https://huggingface.co/mistralai/Mistral-7B-v0.1}{\texttt{mistralai/Mistral-7B-v0.1}} & apache-2.0 \\
        Mistral 7B Chat & \href{https://huggingface.co/mistralai/Mistral-7B-Instruct-v0.2}{\texttt{mistralai/Mistral-7B-Instruct-v0.2}} & apache-2.0 \\
        Llama-3 8B Chat & \href{https://huggingface.co/meta-llama/Meta-Llama-3-8B-Instruct}{\texttt{meta-llama/Meta-Llama-3-8B-Instruct}} & llama3 \\
        Llama-3 70B Chat & \href{https://huggingface.co/meta-llama/Meta-Llama-3-70B-Instruct}{\texttt{meta-llama/Meta-Llama-3-70B-Instruct}} & llama3 \\
        Llama-3.1 8B Chat & \href{https://huggingface.co/meta-llama/Meta-Llama-3.1-8B-Instruct}{\texttt{meta-llama/Meta-Llama-3.1-8B-Instruct}} & llama3.1 \\
        Llama-3.1 70B Chat & \href{https://huggingface.co/meta-llama/Meta-Llama-3.1-70B-Instruct}{\texttt{meta-llama/Meta-Llama-3.1-70B-Instruct}} & llama3.1 \\
        JudgeLM & \href{https://huggingface.co/BAAI/JudgeLM-7B-v1.0}{\texttt{BAAI/JudgeLM-7B-v1.0}} & Non-commercial license \\
        GPT-4 Turbo & \href{https://platform.openai.com/docs/models/gpt-4-turbo-and-gpt-4}{\texttt{gpt-4-turbo-2024-04-09}} & N/A \\
        \bottomrule
    \end{tabular}
\label{tab:assets}
        \captionsetup{skip=8pt} 
\caption{Version and license details for the different models and datasets used in experiments.}
\end{table*}

\section{Model evaluation prompt templates}\label{app:prompt-templates}

\begin{figure*}[htbp]
    \centering
    \centering
    \includegraphics[width=0.8\linewidth]{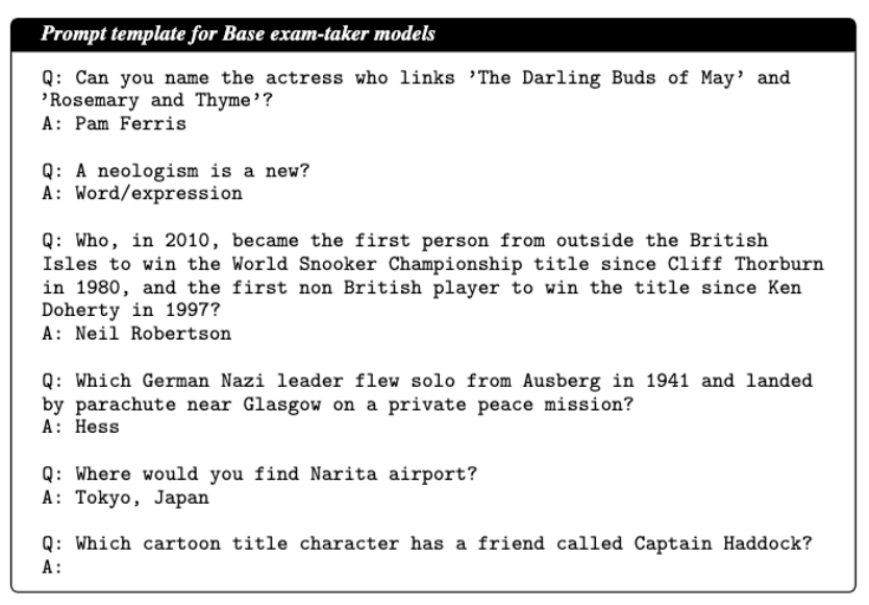}
    \caption{Prompt template for base \evaluatormodels}
    \label{app:template_pretrained}
\end{figure*}

\begin{figure*}[htbp]
    \centering
    \centering
    \includegraphics[width=0.8\linewidth, height=0.5\textheight]{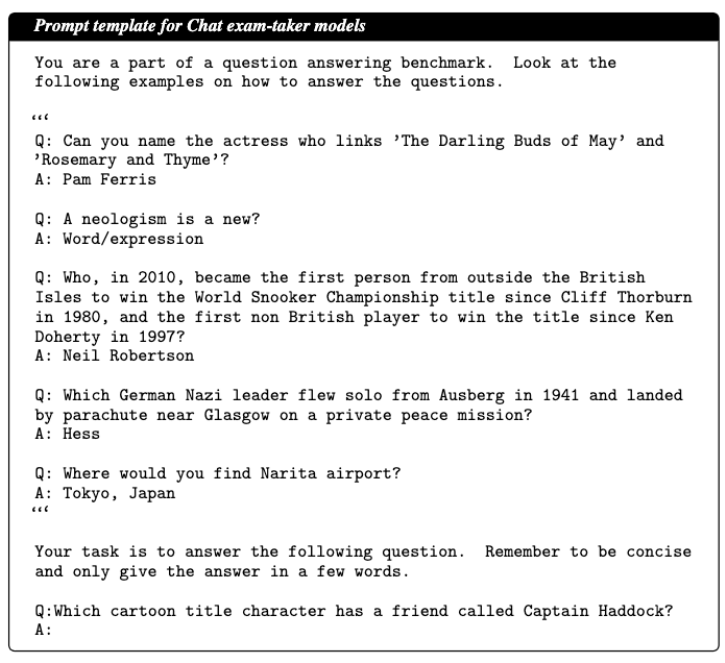}
    \caption{Prompt template for Chat \evaluatormodels}
    \label{app:template_finetuned}
\end{figure*}

In \cref{app:template_pretrained} and \cref{app:template_finetuned}, we show the prompt templates used for the base and chat \evaluatormodels during the question answering process.


\section{Judge LLM Prompt templates}\label{app:judge-prompt-template}
\begin{figure*}[h]
    \centering
        \includegraphics[width=0.8\textwidth, ]{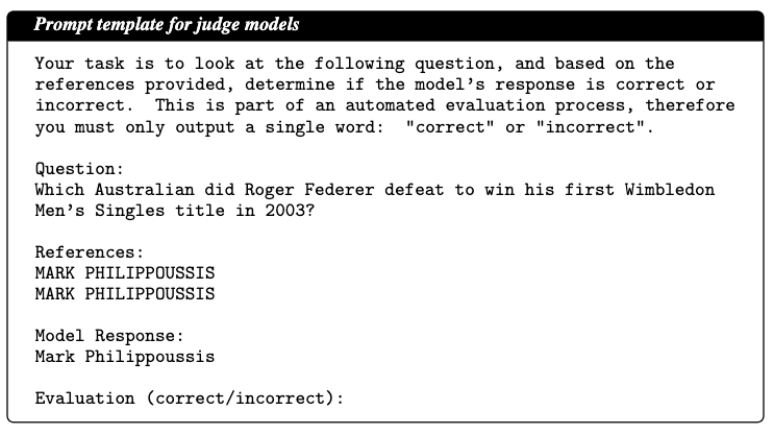}
    \caption{Prompt templates for the \judgemodels}
    \label{app:WithoutGuidelines}
\end{figure*}

In \cref{app:WithoutGuidelines}, we show the prompt template used to guide the \judgemodels during the evaluation process of a 400-question sample from the TriviaQA unfiltered dataset.

\section{Metrics for \judgemodels}
\label{app:metrics}

If one of the annotators is taken to be the reference, then the annotations of the other annotator can be categorized as true positives, false positives, true negatives, and false negatives, with the total number of each of them in a benchmark being represented by $T_P, F_P, T_N,$ and $F_N$ respectively.

\textbf{Percent agreement} is simply the ratio of the numbers of times two annotators agree with each other relative to the total number of annotations. This ratio can have values between $0$ and $1$. For the binary case, the alignment ratio $\rho$ 
is given as

\begin{equation}
    \rho = \frac{T_P + T_N}{T_P + F_P + T_N + F_N}.
\end{equation}

\textbf{Scott's Pi}, \citep{scott1995scottspi}, measures the alignment of two annotators while also taking into account the possibility of agreement by pure chance. This coefficient usually has values above $0$ in most real-world situations. The value of Scott's Pi is given below where $p_o$ is the relative observed agreement, and $p_e$ is the hypothetical probability of chance agreement. 

\begin{figure*}[t]
\begin{equation*}
\pi = \frac{p_o - p_e}{1 - p_e} \nonumber \\
    = \frac{
        \left(\frac{TP + TN}{TP + TN + FP + FN}\right) - 
        \left(\frac{(TP + FP)(TP + FN) + (TN + FN)(TN + FP)}
        {(TP + TN + FP + FN)^2}\right)
    }{
        1 - 
        \left(\frac{(TP + FP)(TP + FN) + (TN + FN)(TN + FP)}
        {(TP + TN + FP + FN)^2}\right)
    }
\end{equation*}
\vspace{-1em}
\label{app:scott-pi}
\end{figure*}

\clearpage
\onecolumn
\section{Human Annotation Guidelines}
\label{app:human_annotation_guidelines}

\mytcolorbox{Human annotation guidelines}{You will be given a question, a set of reference answers and the answer given by an LLM. Your task is to judge if the answer given by the LLM is correct, as if you were the LLMs teacher grading their exam. An answer should be counted as correct if it is semantically equivalent to (one of the) reference answers. In doing so, please follow the following guidelines:}{\begin{itemize}
    \item Underspecified answers (e.g. "December" instead of "December 20") should be marked \textit{incorrect}.
    \item Answers that have more information than requested (e.g. "December 20, in Paris" instead of "December 20") should be marked correct, provided the extra information is not incorrect or contrasting the rest of the answer.
    \item Answers with unnecessary verbosity but correct answers should be marked correct (E.g. ``Thanks for asking this question! The correct answer is: ...").
\end{itemize}
If you have trouble judging whether the answer is correct, for instance because you feel you are lacking knowledge required to judge so, please indicate so by marking the answer "maybe correct" or ``maybe incorrect", so that we can further review it.}

\twocolumn
Preliminary research involved iterative refinement of human annotation guidelines to ensure consistency and reproducibility across annotators with general English semantic knowledge. CS graduate students served as annotators for this experiment. We provide the guidelines used for human evaluation below.

\section{Experiment costs}\label{app:experiment-costs}

The costs for the different experiments described in this work belong in three categories -- GPU-hours for running open-source models on one or more \texttt{Nvidia A100} GPUs, OpenAI credits for making API calls to OpenAI models,\footnote{Pricing details for OpenAI models are available at \url{https://openai.com/api/pricing/}} and human hours for manual annotations of benchmark responses. 
The estimated costs for the final reported experiments are given in \cref{tab:experiment-costs}. 
In addition to this, previous unreported experiments and trials had an approximate cost of 120 GPU-hours, 100 USD in OpenAI credits, and 50 human hours, bringing the total experimental cost for this work to approximately 200 GPU-hours, USD 125 OpenAI credits, and 75 human annotation hours.

\section{Statistical reliability of Evaluation sample} \label{app:downsamplingstddev}

Due to computational constraints discussed in \cref{app:limitations} and \cref{app:experiment-costs}, we limit our evaluation set to randomly sampled 400 questions from TriviaQA \citep{joshi2017triviaqa}. In this section, we further take 5 samples of 300 randomly selected questions from the evaluation set and calculate the mean and standard deviation of Scott's Pi. From \cref{tab:downsampletab}, it can be observed that even on down-sampled sets, the \scottspi values are similar to \cref{fig:llmalignment_b}. Standard deviation of all the \judgemodels from the mean \scottspi is also minimal, barring \judge{EM} lexical match.  

\begin{table}[H]
    \begin{tabular}{lcc}
        \toprule
        Judge Model & Mean \scottspi & Std Dev \\
        \midrule
        Llama3-70B & 0.88 & 0.0046 \\
        Llama3.1-70B & 0.88 & 0.0039 \\
        Llama3.1-8B & 0.78 & 0.0050 \\
        Llama2-13B & 0.75 & 0.0043 \\
        Llama2-70B & 0.69 & 0.0114 \\
        Mistral-7B & 0.67 & 0.0108 \\
        JudgeLM-7B & 0.66 & 0.0026 \\
        Contains & 0.64 & 0.0087 \\
        Llama3-8B & 0.60 & 0.0126 \\
        Llama2-7B & 0.47 & 0.0112 \\
        EM & 0.47  & 0.29 \\
        Gemma-2B & 0.26 & 0.007 \\
        \bottomrule
    \end{tabular}
     \label{tab:downsampletab}
    \centering \captionsetup{skip=8pt} 
     \caption{Weak \scottspi variation for the 5 down-sampled sets indicating robustness for the evaluation sample}
\end{table}

\section{Judge Scores}
\label{app:all_scores}

We show the scores assigned by each \judgemodel to each \evaluatormodel, visualised in \cref{fig:llmalignment_a} in \cref{tab:eval-scores}.



\section{\Evaluatormodel base vs chat analysis}
\label{app:BaseVsChatSupp}

\begin{figure}[t]
\centering
\begin{subfigure}{0.6\textwidth}
    \includegraphics[width=\textwidth]{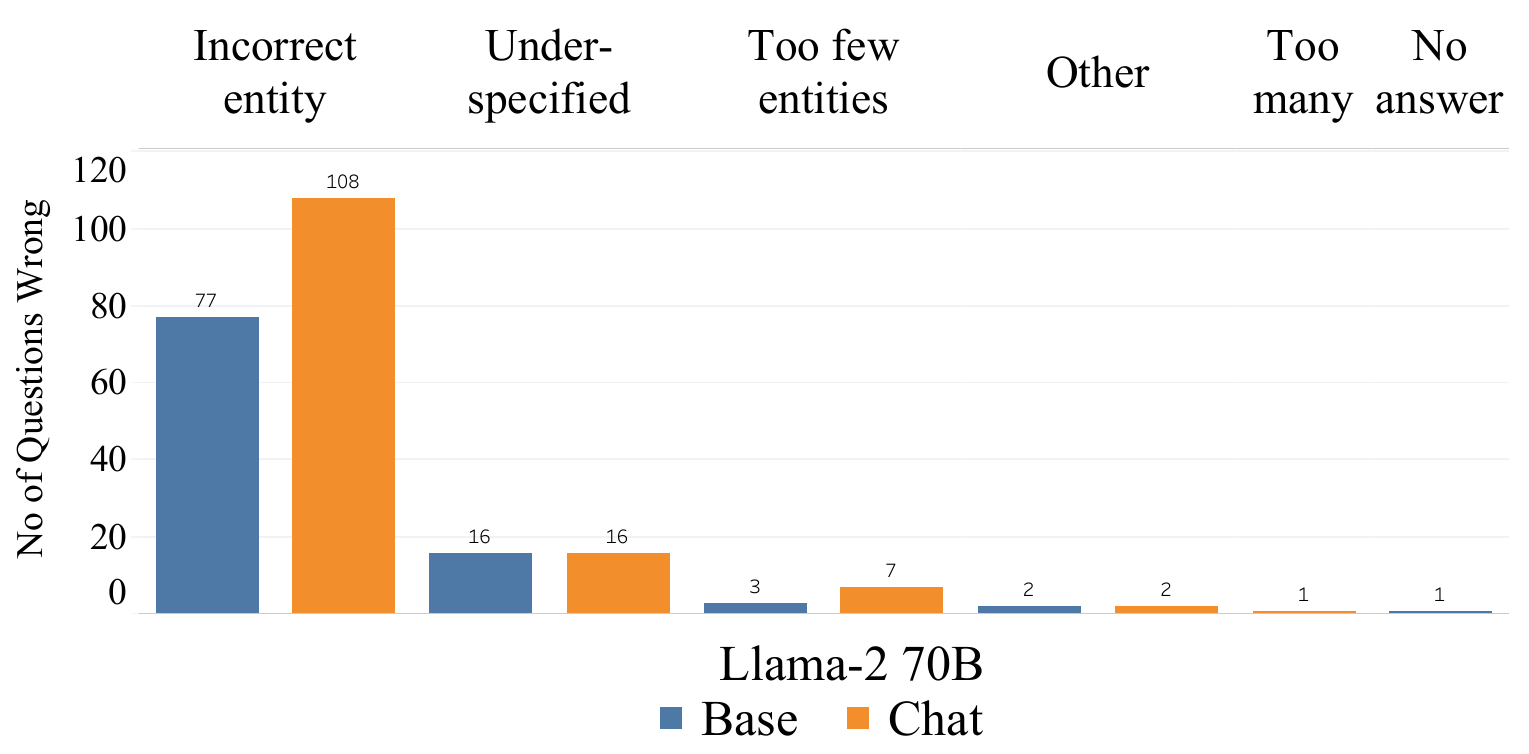}
    \caption{}
     \label{fig:comparisonBarplot}
\end{subfigure}
\hfill
\begin{subfigure}{0.39\textwidth}
\includegraphics[width=\textwidth]{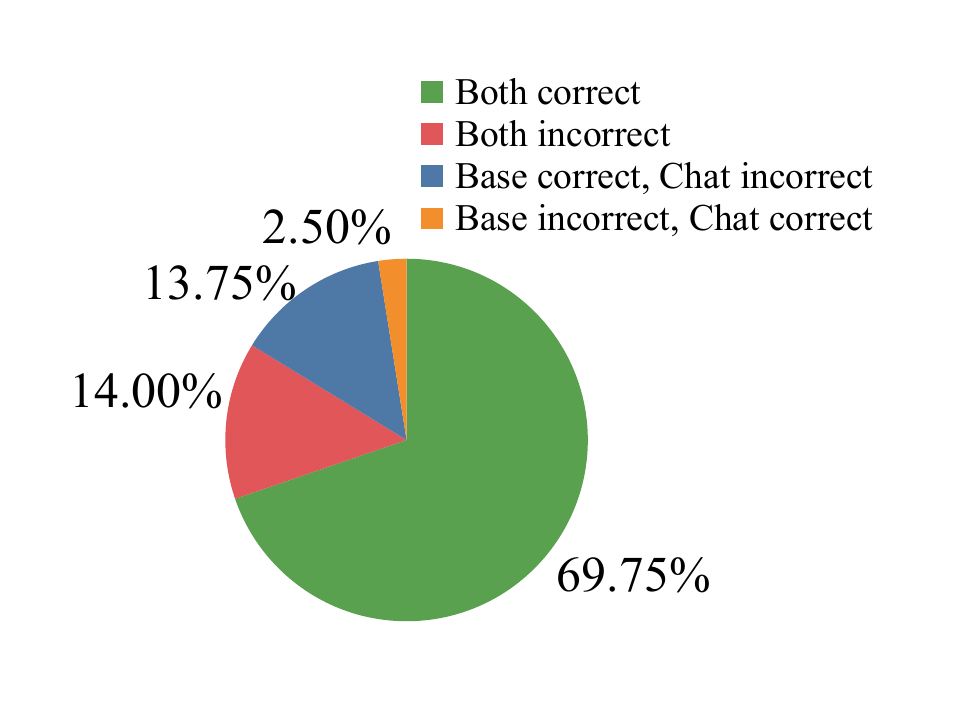}
\caption{}
\label{fig:BaseChatPieChart}
\end{subfigure}
\caption{a) Distribution of incorrect question counts by error codes for \eval{Llama2 70B} Base vs Chat \evaluatormodels evaluated on 400 questions. b) Pie chart showing the percentage of questions categorized by the judgment from Base and Chat models.}
\end{figure}

Given the human judgments we have available, we take the opportunity to investigate the performance differences between base and their corresponding chat models.
In \cref{tab:ScoresBaseChat}, we show the scores assigned by various \judgemodels to four base-chat pairs.
According to the default metric \judge{EM}, the base models outperform the chat models by a large margin.
Interestingly, while this difference gets smaller when the answers are judged by humans (second column) or \judge{GPT-4 Turbo}, there is still a substantial difference for all four pairs, suggesting that the difference is not merely an effect of the increased verbosity of the chat models.
Further evidence for that hypothesis is provided by \cref{fig:BaseChatPieChart}, in which we can see that while 14\% of the errors are shared between the base-chat pairs, almost another 14\% of the examples get judged correctly by the base models but not by the chat models, while the opposite happens in only 2.5\% of the cases.

\onecolumn
\begin{table}[H]
    \centering
    \begin{tabular}{lccc}
        \toprule
        Experiment & GPU-hours & OpenAI credits & Human hours \\
        \midrule
        Main benchmarks & 5 & 2 & - \\
        Main evaluations & 30 & 8 & 10 \\
        Human alignment & 2 & - & 9 \\
        Error analysis & 1.5 & - & 5 \\
        Controlled responses & 15 & - & - \\
        Leniency bias & 5 & 5 & - \\
        Guideline bias & 10 & 5 & 1 \\
        Reference bias & 5 & 4 & 1 \\
        \midrule
        \textbf{Total} & \textbf{73.5} & \textbf{24} & \textbf{26} \\
        \bottomrule
    \end{tabular}
    \label{tab:experiment-costs}
         \captionsetup{skip=8pt} 
     \caption{Estimated costs for the final reported experiments. GPU-hours are in equivalent \texttt{Nvidia A100} hours, OpenAI credits are in USD, and human hours are time spent in manual annotation.}
\end{table}

\begin{table}[h]
\label{tab:eval-scores}
\centering
    \setlength{\tabcolsep}{5pt}
    \begin{tabular}{ccccccccccc}
    &  \multicolumn{9}{c}{\textbf{Exam taker models}} \\
    \cmidrule{2-10}
    & \multicolumn{6}{c}{Llama2} & \multicolumn{2}{c}{Mistral} & GPT-4 \\ 
    & \multicolumn{3}{c}{Base} & \multicolumn{3}{c}{Chat} & Base & Instruct \\
\textbf{Judge Models} & 7B  & 13B & 70B & 7B & 13B & 70B & \multicolumn{2}{c}{7B} \\
\cmidrule[1pt]{2-10}
Llama 3.1 8B & 65.25 & 75.00 & 83.50 & 60.25 & 70.50 & 75.50 & 73.75 & 59.00 & \textbf{89.00} \\ 
Llama 3.1 70B & 62.00 & 74.25 & 85.00 & 55.50 & 64.75 & 74.00 & 72.25 & 60.50 & \textbf{92.25} \\ \midrule
Llama 3 8B & 76.00 & 83.25 & 91.50 & 73.25 & 82.75 & 85.25 & 81.75 & 76.0 & \textbf{97.25} \\ 
Llama 3 70B & 64.25 & 75.50 & 86.50 & 57.00 & 64.00 & 75.75 & 73.5 & 62.50 & \textbf{92.75} \\ \midrule
Llama 2 7B & 80.50 & 85.25 & 92.00 & 80.50 & 70.75 & 90.75 & 84.00 & 83.25 & \textbf{97.75} \\ 
Llama 2 13B & 68.25 & 75.50 & 86.50 & 63.25 & 62.75 & 77.50 & 74.50 & 67.50 & \textbf{93.5} \\ 
Llama 2 70B & 71.25 & 80.5 & 90.25 & 67.50 & 74.75 & 81.25 & 80.0 & 72.5 & \textbf{96.75} \\ \midrule
Mistral 7B & 72.50 & 80.75 & 90.50 & 69.00 & 74.75 & 82.50 & 80.25 & 72.00 & \textbf{96.25} \\ \midrule
Gemma 2B & 79.75 & 87.00 & \textbf{91.25} & 58.50 & 41 & 68.50 & 84.0 & 55.75 & 80.50 \\ \midrule
JudgeLM & 69.50 & 77.75 & 86.25 & 63.75 & 48.0 & 82.75 & 77.25 & 71.0 & \textbf{94.50} \\ \midrule
GPT-4 & 60.50 & 71.50 & 82.50 & 54.50 & 59.0 & 73.0 & 69.75 & 56.50 & \textbf{90.0} \\ \midrule
Exact Match & 46.75 & 56.00 & \textbf{63.75} & 24.00 & 0.25 & 36.25 & 59.50 & 20.25 & 58.25 \\ 
Contains Match & 50.75 & 60.00 & 68.00 & 39.00 & 46.25 & 59.50 & 57.25 & 44.00 & \textbf{70.00} \\ \midrule
Human Eval & 62.50 & 72.75 & 83.75 & 56.00 & 56.50 & 72.25 & 71.75 & 60.75 & \textbf{91.50} \\
\bottomrule
\end{tabular}
 \captionsetup{skip=8pt} 
\caption{\Judgemodel score card for every \evaluatormodel.}
\end{table}

\twocolumn


\begin{table}[b]
    \centering
\label{tab:ScoresBaseChat}
 \captionsetup{skip=8pt} 
\caption{Scores of base and chat models by various judges}
\setlength{\tabcolsep}{6pt}
\begin{tabular}{p{2.5cm}cccccccccc}
\toprule
& \multicolumn{10}{c}{\textbf{Judge models}} \\ \cmidrule(lr){2-11}
    \makecell{Base-Chat\\ pair} & \multicolumn{2}{c}{EM} & \multicolumn{2}{c}{Contains} & \multicolumn{2}{c}{Human} & \multicolumn{2}{c}{\makecell{GPT-4\\Turbo}} & \multicolumn{2}{c}{\makecell{Llama-3\\70B}} \\
    \cmidrule{2-11}
    & Base & Chat & Base & Chat & Base & Chat & Base & Chat & Base & Chat \\
    \makecell{Llama-2 7B} & \textbf{46.75} & 24.00 & \textbf{50.75} & 39.00 & \textbf{62.25} & 56.00 & \textbf{60.50} & 54.50 & \textbf{64.25} & 57.00\\
    \makecell{Mistral 7B} & \textbf{59.50} & 20.25 & \textbf{57.25} & 44.00 & \textbf{71.75} & 60.75 & \textbf{69.75} & 56.50 & \textbf{73.50} & 62.50\\
    \makecell{Llama-2 13B} & \textbf{ 56.00} & 0.25 & \textbf{60.00} & 46.25 & \textbf{72.75} & 56.50 & \textbf{75.00} & 59.00 & \textbf{76.50} & 64.00\\
    \makecell{Llama-2 70B} & \textbf{63.75} & 36.25 &  \textbf{68.00} & 59.50 & \textbf{83.75} & 72.25 & \textbf{82.50} & 73.00 & \textbf{86.50} & 75.75\\
\bottomrule
\end{tabular}
\end{table}

We consider two alternative hypotheses:\begin{itemize}\setlength\itemsep{0.1em}
    \item[i)] The chat models have a worse understanding of the particular prompt format, which is tuned more to fit base models; or
    \item[ii)] The chat models have `unlearned' some knowledge during their alignment training.
\end{itemize}

To disentangle these two factors, we manually analyse 400 questions for \eval{Llama-2 70B} and \eval{Llama-2 70B-chat}, using our earlier error codes.
The results, shown in \cref{fig:comparisonBarplot}, sugest that, at least to some extent, the difference between base and chat models is in fact due to `unlearning' of knowledge: while the number of errors is more or less equal among most categories, there is a stark difference in the \emph{incorrect entity} category.
Substantially more often than the base models, the chat models do answer the question with a semantically plausible but incorrect entity.
In \cref{tab:KnowledgeUnlearningExample1}-\cref{tab:KnowledgeUnlearningExample3}, we provide examples of such cases.
The results do not show any evidence to support the first hypothesis: the number of errors where the answer cannot be parsed or is just entirely incorrect does not differ between base and chat models.

\section{\Evaluatormodel ranking correlation }
\label{app:correlationcoefftable}

In \cref{tab:judges_rho_reversed}, We use the Spearman Rank correlation coefficient  \citep{spearman1904spearman} to assess the rankings of the \evaluatormodels. To validate these rankings, we randomly select 6 out of \nexamtakers \evaluatormodels across 5 samples, subsequently calculating the mean ($\rho$) and standard deviation ($\sigma$) of the rankings. The results reveal that the \judge{contains} model exhibits the highest stability and $\rho$ among the rankings, while the majority of judge models achieve a coefficient exceeding 0.7, indicating a strong alignment. Notably, smaller models such as \judge{Mistral 7B} perform on par with \judge{\gpt}, highlighting the robustness of smaller models in maintaining rankings.

\begin{table}[H]
\label{tab:judges_rho_reversed}
    \centering
    \begin{tabular}{lcc}
        \toprule
        Judges & $\rho$ & $\sigma$ \\
        \midrule
        Contains & 0.99 & 0.02 \\
        Mistral-7B & 0.98 & 0.03 \\
        GPT-4 & 0.98 & 0.03 \\
        Llama2-13B & 0.95 & 0.18 \\
        JudgeLM-7B & 0.95 & 0.05 \\
        Llama2-7B & 0.94 & 0.04 \\
        Llama3.1-70B & 0.94 & 0.07 \\
        Llama3-70B & 0.93 & 0.05 \\
        Llama3.1-8B & 0.89 & 0.10 \\
        Llama3-8B & 0.86 & 0.07 \\
        Llama2-70B & 0.84 & 0.13 \\
        Gemma-2B & 0.71 & 0.20 \\
        EM & 0.67 & 0.13 \\
        \bottomrule
    \end{tabular}
            \captionsetup{skip=5pt}
    \caption{Spearman Rank Correlation Coefficient $\rho$.}
\end{table}

\section{Too much info confuses judges}
\label{app:TMI}
\begin{figure}[H]
    \centering
    \resizebox{0.8\textwidth}{!}{
        \includegraphics{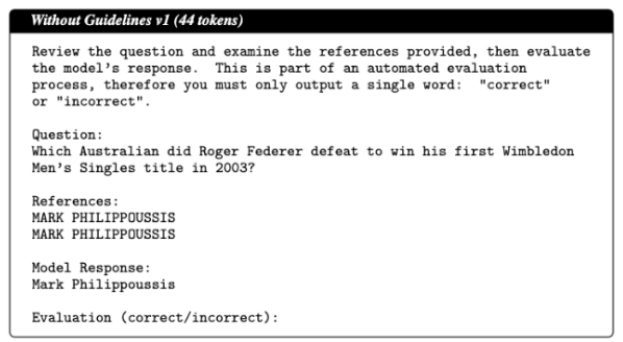}
    }
    \caption{\textit{Without Guidelines v1} prompt template for the \judgemodels}
    \label{app:WithoutGuidelines_v1}
\end{figure}

\begin{figure}[t]
    \centering
    \resizebox{0.8\textwidth}{!}{
        \includegraphics{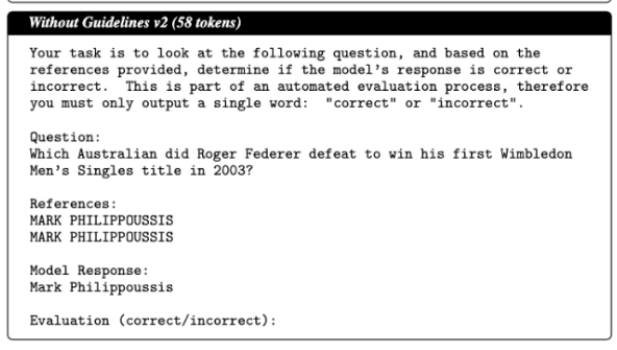}
    }
    \caption{\textit{Without Guidelines v2} prompt template for the \judgemodels}
    \label{app:WithoutGuidelines_v2}
\end{figure}

\begin{figure}[h]
    \centering
    \resizebox{0.7\textwidth}{!}{
        \includegraphics{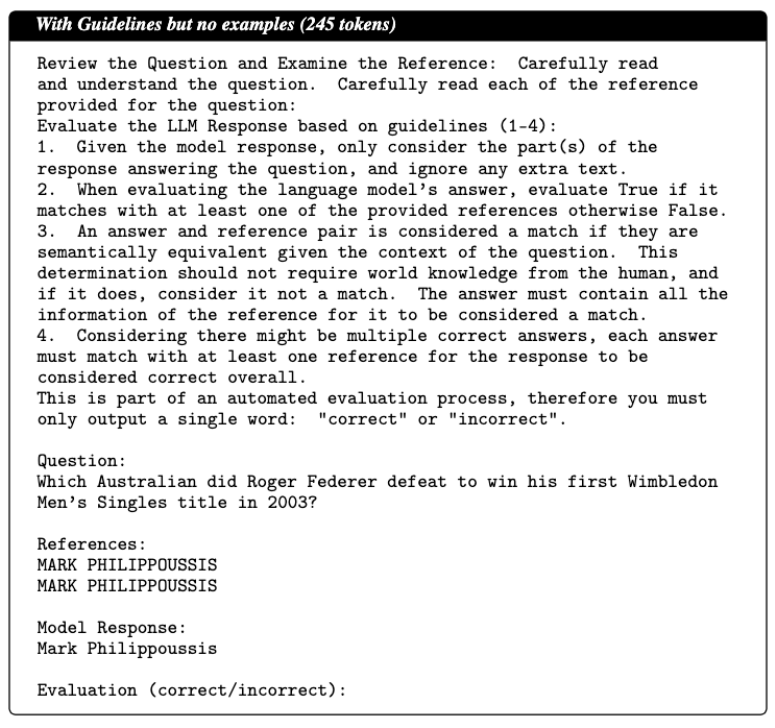}
    }
    \caption{\textit{Guidelines without examples} Prompt template for the \judgemodels}
    \label{app:GuidelinesWithoutExamples}
\end{figure}

\begin{figure}[ht]
    \centering
    \resizebox{0.7\textwidth}{!}{
        \includegraphics{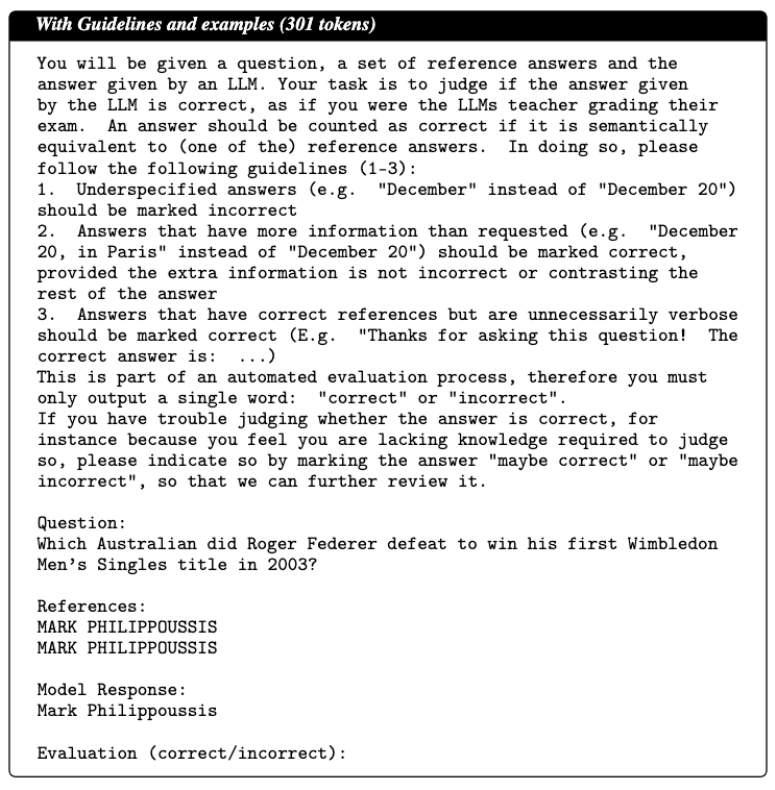}
    }
    \caption{\textit{Guidelines with Examples} Prompt template for the \judgemodels}
    \label{app:GuidelinesWithExamples}
\end{figure}

In \cref{app:WithoutGuidelines_v1}-\ref{app:GuidelinesWithExamples}, we report the guidelines we used for the experiments in \cref{sec:analysis:subsec:instructions}. 
The simplest prompt used is \textit{Without Guidelines v1} (see \cref{app:WithoutGuidelines_v1}) where we define a sequential and structured process for the \judgemodel. 
In \textit{Without Guidelines v2} (see \cref{app:WithoutGuidelines_v2}), we add an additional focus on the overall task and outcome as well. 
For \textit{Guidelines without examples} (see \cref{app:GuidelinesWithoutExamples}), we provide the \judgemodels with detailed instructions about the task at hand, along with explicit guidelines on how to evaluate the answers. 
Additionally, for \textit{Guidelines with examples}(see \cref{app:GuidelinesWithExamples}), we also provide examples to the \judgemodels for further reference.

\onecolumn

\definecolor{darkgreen}{rgb}{0.0, 0.5, 0.0}
\begin{table}[H]
\centering
\label{tab:KnowledgeUnlearningExample1}
\begin{tabular}{|>{\raggedright\arraybackslash}m{2.5cm}|>{\raggedright\arraybackslash}m{10cm}|}
\hline
\multicolumn{2}{|c|}{\textbf{Question:}} \\
\multicolumn{2}{|c|}{\texttt{Which British artist's works include `The First Real Target'?}} \\
\hline
\textbf{References} & \rule{0pt}{3ex}\texttt{Peter Blake, Peter Balke, Sir Peter Blake}\rule[-1ex]{0pt}{1ex} \\
\hline
\textbf{LLama-2 70B Base} & \rule{0pt}{3ex}\textcolor{darkgreen}{\texttt{Peter Blake}}\rule[-1ex]{0pt}{1ex} \\
\hline
\textbf{LLama-2 70B Chat} & \rule{0pt}{3ex}\textcolor{red}{\texttt{Patrick Caulfield}}\rule[-1ex]{0pt}{1ex} \\
\hline
\textbf{Mistral 7B Base} & \rule{0pt}{3ex}\textcolor{red}{\texttt{David Hockney}}\rule[-1ex]{0pt}{1ex} \\
\hline
\textbf{Mistral 7B Chat} & \rule{0pt}{3ex}\textcolor{red}{\texttt{Damien Hirst}}\rule[-1ex]{0pt}{1ex} \\
\hline
\end{tabular}
\label{tab:KnowledgeUnlearningExample1}
\captionsetup{skip=5pt}
\caption{Knowledge unlearning example 1.}
\end{table}

\begin{table}[H]
\label{tab:KnowledgeUnlearningExample2}
\centering
\begin{tabular}{|>{\raggedright\arraybackslash}m{2.5cm}|>{\raggedright\arraybackslash}m{10cm}|}
\hline
\multicolumn{2}{|c|}{\textbf{Question:}} \\
\multicolumn{2}{|c|}{\texttt{Who was the first cricketer to score 10,000 test runs?}} \\
\hline
\textbf{References} & \rule{0pt}{3ex}\texttt{Sunil Gavaskar, Sunil Manohar Gavaskar, SM Gavaskar, Sunny gavaskar, Gavaskar}\rule[-1ex]{0pt}{1ex} \\
\hline
\textbf{LLama-2 70B Base} & \rule{0pt}{3ex}\textcolor{darkgreen}{\texttt{Sunil Gavaskar}}\rule[-1ex]{0pt}{1ex} \\
\hline
\textbf{LLama-2 70B Chat} & \rule{0pt}{3ex}\textcolor{red}{\texttt{Sachin Tendulkar}}\rule[-1ex]{0pt}{1ex} \\
\hline
\textbf{Mistral 7B Base} & \rule{0pt}{3ex}\textcolor{red}{\texttt{Sachin Tendulkar}}\rule[-1ex]{0pt}{1ex} \\
\hline
\textbf{Mistral 7B Chat} & \rule{0pt}{3ex}\textcolor{red}{\texttt{Sachin Tendulkar}} \texttt{was the first cricketer to score 10,000 runs in Test matches.}\rule[-1ex]{0pt}{1ex} \\
\hline
\end{tabular}
\label{tab:KnowledgeUnlearningExample2}
\captionsetup{skip=5pt}
\caption{Knowledge unlearning example 2}
\end{table}

\begin{table}[H]
\label{tab:KnowledgeUnlearningExample3}
\centering
\begin{tabular}{|>{\raggedright\arraybackslash}p{2.5cm}|>{\raggedright\arraybackslash}p{10cm}|}
\hline
\multicolumn{2}{|c|}{\textbf{Question:}} \\
\multicolumn{2}{|c|}{\parbox{12cm}{\texttt{`Uncle Harry's Coat' was the first garment produced by which famous jacket manufacturer, based in Simonside, Newcastle Upon Tyne?}}} \\
\hline
\textbf{References} & \rule{0pt}{3ex}\texttt{Barbour}\rule[-1ex]{0pt}{1ex} \\
\hline
\textbf{LLama-2 70B Base} & \rule{0pt}{3ex}\textcolor{darkgreen}{\texttt{Barbour}}\rule[-1ex]{0pt}{1ex} \\
\hline
\textbf{LLama-2 70B Chat} & \rule{0pt}{3ex}\textcolor{darkgreen}{\texttt{Barbour}}\rule[-1ex]{0pt}{1ex} \\
\hline
\textbf{Mistral 7B Base} & \rule{0pt}{3ex}\textcolor{darkgreen}{\texttt{Barbour}}\rule[-1ex]{0pt}{1ex} \\
\hline
\textbf{Mistral 7B Chat} & \rule{0pt}{3ex}\textcolor{red}{\texttt{Jack Walker \& Sons}}\rule[-1ex]{0pt}{1ex} \\
\hline
\end{tabular}
\label{tab:KnowledgeUnlearningExample3}
\captionsetup{skip=5pt}
\caption{Knowledge unlearning example 3}
\end{table}

\clearpage

\begin{figure}[htbp]
    \centering
    \resizebox{0.6\textwidth}{!}{
        \includegraphics{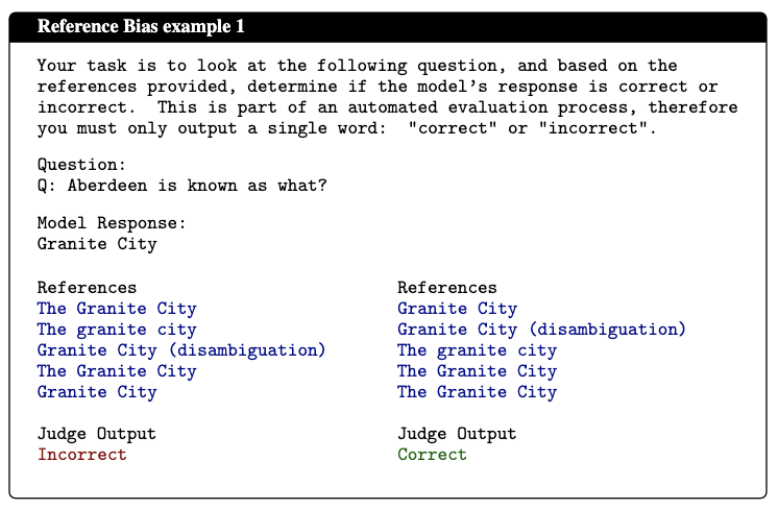}
    }
    \caption{Example of \judge{Llama2-7B} getting confused when the order of the references are changed}
    \label{app:ReferenceBiasExample1}
\end{figure}

\begin{figure}[H]
    \centering
    \resizebox{0.6\textwidth}{!}{
        \includegraphics{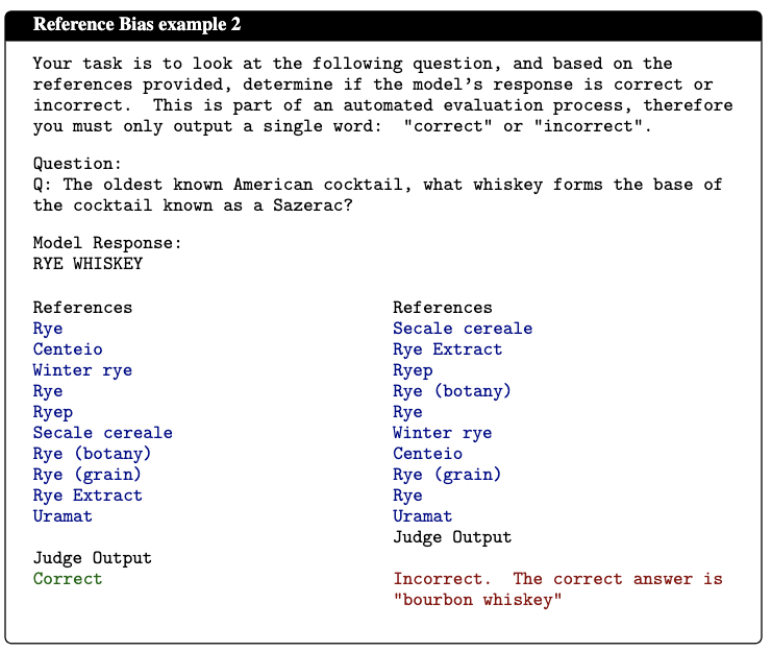}
    }
    \caption{Example of \judge{Llama2-7B} failing to identify the task by changing the order of the references.}
    \label{app:RefernceBiasExample2}
\end{figure}

\clearpage
\twocolumn
\section{\Judgemodels are sensitive to reference order}
\label{app:ref-bias-exp}

We investigate the judges' sensitivity to reference order by providing the same prompt, question and model response to the \judgemodels, but shuffling the reference order in three different permutations. We compute the consistency score of the model as the percentage of questions for which it gives the same judgment all the 3 times. 
%
We observe that the model is more likely to evaluate an answer as correct if the corresponding reference appears early in the list of references (see \cref{app:ReferenceBiasExample1}).
%
The smaller \judgemodels sometimes fail to capture all the information in the prompt, and provide judgement based on their own knowledge rather than going by the references (see  \cref{app:RefernceBiasExample2}).


\section{Leniency Bias}\label{app:leniency-bias}

As described in \cref{sec:leniency-bias}, for the purpose of the leniency bias experiments, we assume that a judge assigns the correct judgment with a probability of $P_c$ and randomly assigns the rest of the samples to be \texttt{“correct”} with a probability $P_+$.
In this section, we derive the mathematical expressions for $P_c$ and $P_+$. We assume that in the case of misalignment between the evaluation criteria of guidelines and \judgemodels, the probability of getting an evaluation of \texttt{``correct''} is independent of the actual correctness of the answer (i.e.\ the \judgemodel effectively flips a coin to give out its judgement). For any given benchmark and \judgemodel, we denote the ground-truth score as $s$, and the true positive and true negative rates as $t_P$ and $t_N$, respectively, all normalized to be between $0$ and $1$.

Now, based on our assumptions, the true positives, where the \evaluatormodel response is correct, and also correctly identified by the \judgemodel to be correct, would be comprised of two possible cases: 1) The judge evaluates it correctly according to the given evaluation criteria with a probability of $P_c$; and 2) The judge does not evaluate it according to the given criteria with a probability of $1-P_c$, but the evaluation still happens to be correct with a probability of $P_+$. With the total ratio of the correct responses being $s$, the true positive rate is therefore given by --

\begin{equation}\label{eq:tp}
    t_P = s[P_c + (1-P_c)P_+]
\end{equation}

Similarly, the true negatives, where the \evaluatormodel response is incorrect, and also correctly identified by the \judgemodel to be incorrect, would also be comprised of two cases: \textbf{1)} The judge evaluates it correctly according to the given evaluation criteria with a probability of $P_c$.\textbf{2)} The judge does not evaluate it according to the given criteria with a probability of $1-P_c$, but the evaluation still happens to be correct with a probability of $1-P_+$. With the total ratio of the incorrect responses being $1-s$, the true negative rate is therefore given by --

\begin{equation}\label{eq:tn}
    t_N = (1-s)[P_c + (1-P_c)(1-P_+)].
\end{equation}

Using \cref{eq:tn}, we can derive the following. 

\begin{align}
    t_N &= (1-s)[P_c + (1-P_c)(1-P_+)] \\
    &= P_c + 1 - P_+ - P_c + P_cP_+ \\
    &\quad - sP_c  -s + sP_+ + sP_c - sP_cP_+ \\
    &=  1 - P_+ + P_cP_+ -s + sP_+  - sP_cP_+ \\
    &= 1 - s - P_+(1 - P_c - s + sP_c) \\
    &= 1 - s - P_+(1-s)(1-P_c) \\
    \implies P_+ &=\frac{1-s - t_N}{(1-s)(1-P_c)} \\
    &= \frac{1 - \frac{t_N}{1-s}}{1-P_c}
\end{align}

Substituting the value of $P_+$ in \cref{eq:tp}, we get:

\begin{align}
    t_P &= s[P_c + (1-P_c)P_+] \\
    &= s\Bigg[P_c + (1-P_c)\frac{1 - \frac{t_N}{1-s}}{1-P_c}\Bigg] \\
    &= s\bigg[P_c + 1 - \frac{t_N}{1-s}\bigg] \\
    \implies \frac{t_P}{s} &= P_c + 1 - \frac{t_N}{1-s} \\
    \implies P_c &= \frac{t_P}{s} + \frac{t_N}{1-s} - 1
\end{align}

\begin{figure}[H]
\begin{subfigure}[b]{0.45\textwidth}
    \centering
    
    \begin{tabular}{lrrr}
      \toprule
      \Judgemodel & $\pi$ & $P_c$ & $P_+$ \\
      \midrule
          \judge{Gemma-2B} & 0.26 & 0.38 & 0.87 \\
          \judge{Llama2-7B} & 0.47 & 0.63 & 0.75 \\
          \judge{Llama3-8B} & 0.59 & 0.63 & 0.74 \\
          \judge{JudgeLM-7B} & 0.65 & 0.68 & 0.19 \\
          \judge{Mistral-7B} & 0.66 & 0.70 & 0.87 \\
          \judge{Llama2-70B} & 0.69 & 0.66 & 0.99 \\
          \judge{Llama2-13B} & 0.74 & 0.74 & 0.87 \\
          \judge{Llama3.1-8B} & 0.77 & 0.77 & 0.82 \\
           \judge{GPT-4} & 0.87 & 0.87 & 0.69 \\
          \judge{Llama3.1-70B} & 0.88 & 0.88 & 0.82 \\
          \judge{Llama3-70B} & 0.88 & 0.87 & 0.90 \\
      \bottomrule
\end{tabular}
\caption{}
\label{tab:p-vals-full}
\end{subfigure}%
\hspace{1cm}
\begin{subfigure}[t]{0.45\textwidth}
  \centering
  \includegraphics[width=\linewidth]{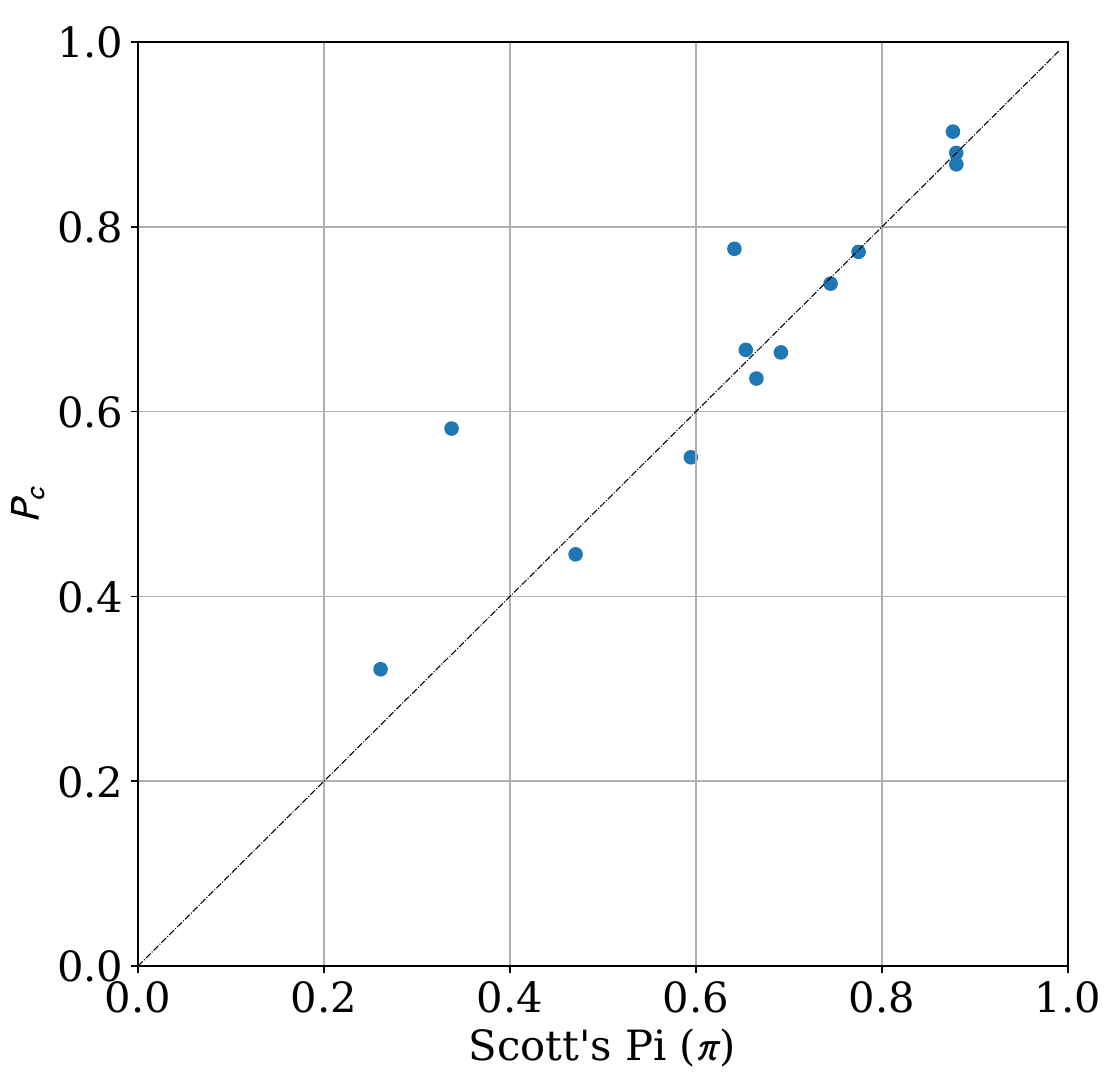}
  \caption{}
  \label{fig:k-p-corr}
\end{subfigure}
\caption{a) Estimated values of $P_c$ and $P_+$ for different \judgemodels. b) Pearson's correlation coefficient between $\pi$ and $P_c$ for \judgemodels.}
\label{fig:leniency-bias-full}
\end{figure}

The values of $P_c$ and $P_+$ can be estimated from observed data using the derived expressions. 
The estimated probabilities using this method, with human evaluation as the reference, are shown in \cref{tab:p-vals-full}.

To validate these derived values, we observe the correlation between the estimated values of $P_c$ and  Scott's Pi ($\pi$). 
As shown in \cref{fig:k-p-corr}, we observe that the estimated values of $P_c$ are highly correlated to the \scottspi values for the \judgemodels, with a Pearson correlation coefficient of $0.98$.

\begin{figure}[t]
    \centering
    \begin{subfigure}[b]{0.4\textwidth}
        \centering
        \includegraphics[width=\textwidth]{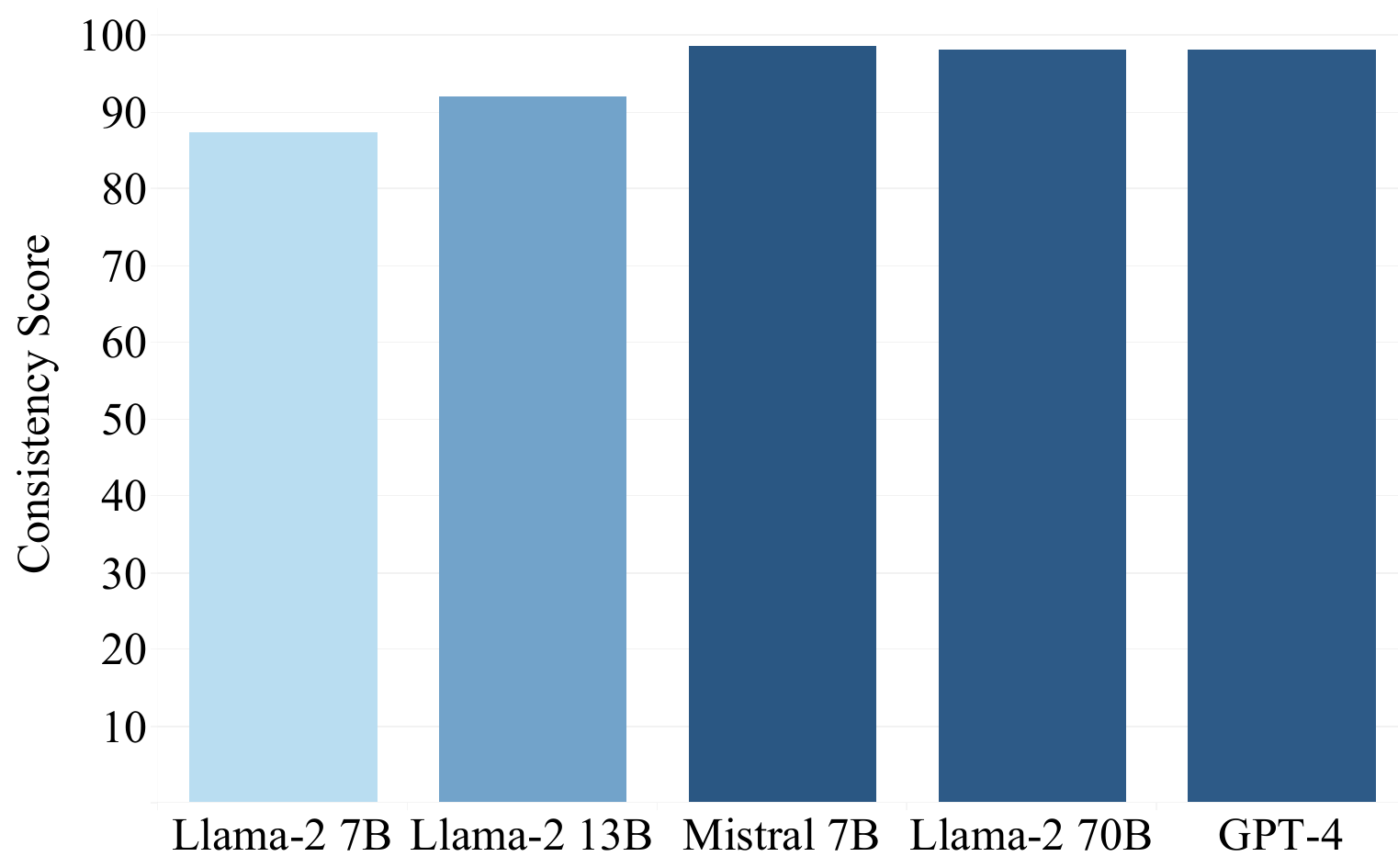}
        \label{fig:consistency}
    \end{subfigure}
    \caption{\textbf{Leniency bias and answer consistency.} Consistency score, defined as the percentage of questions for which the \judgemodel gives the same judgment for three different answer orders.}
\end{figure}



\end{document}